\documentclass[runningheads]{llncs}

\usepackage{eccv}

\usepackage{eccvabbrv}

\usepackage{graphicx}
\usepackage{booktabs}

\usepackage{multirow}   
\usepackage{makecell}  
\usepackage{graphicx}  
\usepackage{algorithm}
\usepackage{algpseudocode}
\usepackage{mathtools}

\usepackage[accsupp]{axessibility}  

\newcommand{\equalcontrib}{\ensuremath{^*}}
\newcommand{\corrauth}{\ensuremath{^\dagger}}

\usepackage{hyperref}

\usepackage{orcidlink}

\makeatletter
\renewcommand\paragraph{\@startsection{paragraph}{4}{\z@}%
  {-12pt \@plus -4pt \@minus -2pt}
  {-0.5em}
  {\normalfont\normalsize\bfseries}} 
\makeatother
\begin{document}

\title{Geometry-Aware Style Transfer in 3D Gaussian Splatting} 



\author{
Min Hyeok Bang\equalcontrib\inst{1} \and
Jun Hyeong Kim\equalcontrib\inst{1} \and
Seung-Wook Kim\corrauth\inst{2}\orcidlink{0000-0002-6004-4086}  \and \\
Se-Ho Lee\corrauth\inst{1}\orcidlink{0000-0003-0366-581X}
}

\authorrunning{M. H.~Bang et al.}

\institute{Department of Computer Science and Artificial Intelligence/Center for Advanced Image Information Technology, Jeonbuk National University, Jeonju-si, South Korea \and
Division of Electronic and Communication Engineering, Pukyong National University, Busan, South Korea \\
\email{\{bunsss77,kjun627\}@jbnu.ac.kr, swkim@pknu.ac.kr, seholee@jbnu.ac.kr}\\
}

\maketitle
\begingroup
\renewcommand{\thefootnote}{}
\footnotetext{
\textsuperscript{*} These authors contributed equally.\\
\textsuperscript{\ensuremath{\dagger}} Corresponding authors.
}
\endgroup

\begin{abstract}
  In this paper, we present a novel geometry-aware style transfer framework for 3D Gaussian splatting (3DGS) that simultaneously transfers appearance attributes and geometric structures. Unlike prior works that primarily focus on color-based stylization and often overlook structural adaptation, our method explicitly incorporates geometry adaptation through a decoupled optimization scheme that alternately updates color and geometry parameters. This strategy alleviates potential interference between color and geometry updates, leading to stable and consistent scene-level geometry transformation. The decoupled optimization is enabled by the proposed geometry-aware contrastive feature matching (GCFM). GCFM integrates RGB, depth, and edge cues into a contrastive objective and is employed in both optimization phases to effectively transfer structural characteristics from style images to Gaussian primitives. Extensive experiments show that our approach achieves superior performance in both qualitative fidelity and quantitative metrics, significantly outperforming existing 3DGS-based stylization methods. Our code is available at \href{https://github.com/oweixx/gast}{https://github.com/oweixx/gast}.

\end{abstract}

\begin{figure*}[t] 
\centering 
\includegraphics[width=\linewidth]{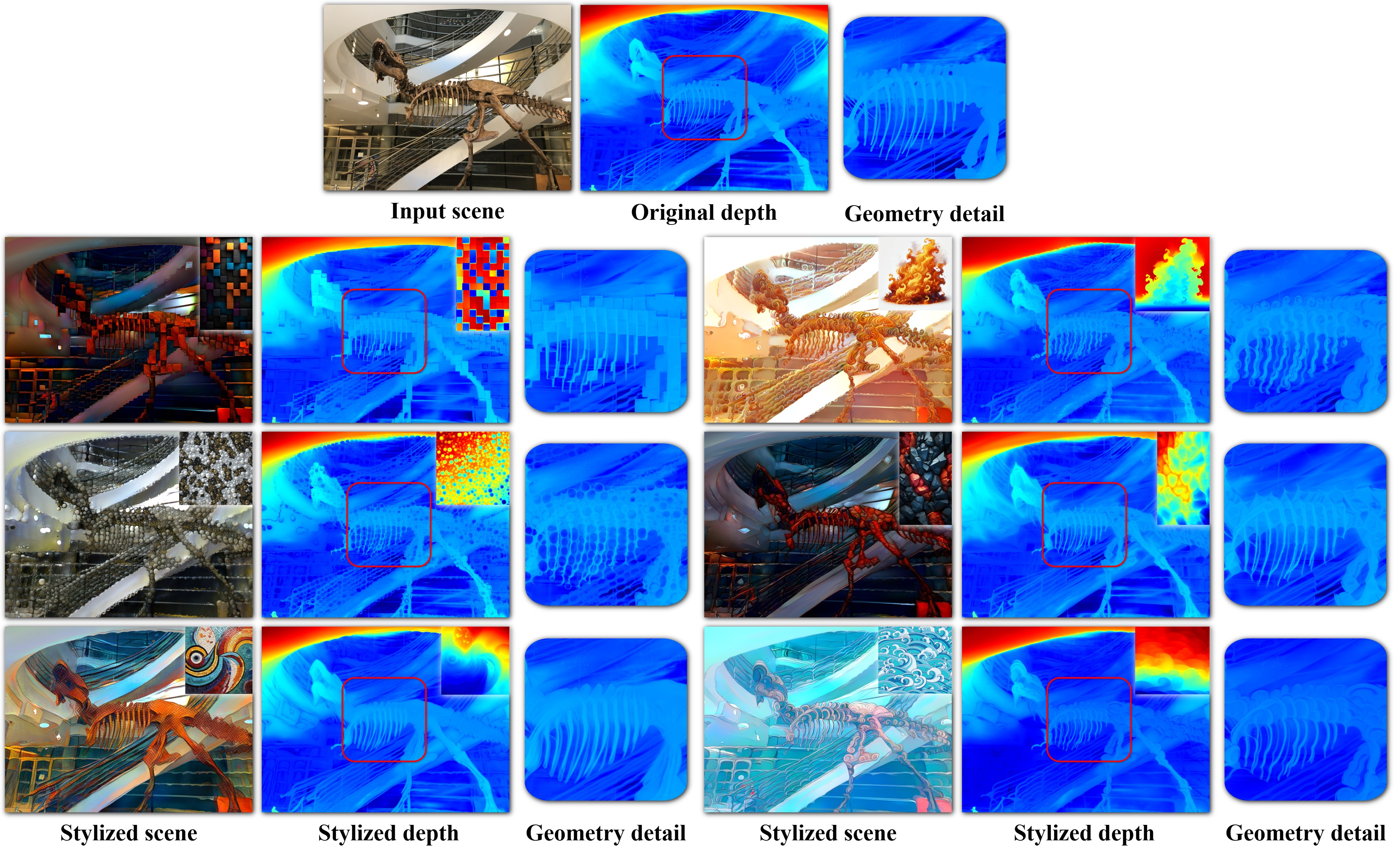} 
\vspace{-7mm}
\caption{Examples of geometry-aware style transfer in 3D Gaussian splatting (3DGS). Given an input scene and various style guides, our method stylizes both color and geometry while preserving overall scene structure. Highlighted depth regions show how geometric details adaptively reflect structural cues from the target style. }
\label{fig:intro}
\vspace{-8mm}
\end{figure*}

\section{Introduction}
\label{sec:intro}

Neural style transfer~\cite{Gatys} sparked extensive research on transferring the artistic characteristics of a style image onto a content image by manipulating color, texture, and high-level feature statistics.
Several works have improved visual quality and user controllability of style transfer using CNNs~\cite{Johnson,Ulyanov}, GANs~\cite{CycleGAN,MUNIT,DRIT}, and diffusion models~\cite{StyleDiffusion}. 
While these approaches achieve impressive results on single-view images, they are inherently limited to 2D representations and cannot maintain consistent stylization across different viewpoints.
With the advent of neural radiance fields (NeRFs)~\cite{NeRF}, some researchers began exploring 3D style transfer to render stylized scenes with a consistent appearance across different viewpoints. NeRF-based methods~\cite{ARF,Geometry,StylizedNeRF,StyleRF,CoARF,SNeRF,refnpr} successfully introduced geometric and structural awareness into style transfer, but they suffer from heavy optimization costs and slow rendering, which limit their practical use in interactive applications.

As a powerful alternative for real-time novel view synthesis, 3D Gaussian splatting~(3DGS)~\cite{3DGS} has been recently proposed. By modeling a scene as a set of anisotropic 3D Gaussians with learnable positions, scales, colors, and opacities, 3DGS achieves both high-quality rendering and remarkable efficiency compared to NeRFs. Its explicit and compact nature has made it attractive for applications beyond photorealistic reconstruction, such as interactive graphics and immersive content creation, leading to active exploration of 3D style transfer~\cite{StyleGaussian,SGSST,GStyle,StylizedGS,ABCGS,clipgaussian}.

However, applying style transfer to 3DGS often introduces a significant challenge: the rendered output is far more sensitive to changes in geometry than to changes in color. While color modifications mainly affect the surface appearance, direct manipulation of geometric parameters alters the underlying structure of a scene, which means that even minor miscalculations in geometry can drastically degrade visual quality. This difficulty explains why most existing style transfer methods tend to adopt a conservative approach. They impose strong constraints on geometry and focus on adapting color-related attributes for stylization, thereby limiting the full expressive potential of 3D style transfer. 
Essentially, such a conservative approach is analogous to applying wood-patterned wallpaper onto a piece of furniture: while the paper conforms to the furniture’s overall shape and surface details, it cannot modify the underlying structure to embody the new style.
As a result, the outcome would be a superficial stylization that lacks the authentic geometric character of real wood—its grooves, indentations, and structural irregularities.
In contrast, our work treats both geometry and appearance as active components in the stylization process. We move beyond merely \textit{painting} a surface and instead \textit{sculpt} the stylistic features by actively shaping both the form and texture.

To this end, we propose a novel geometry-aware style transfer framework for 3DGS that transfers both stylistic attributes and geometric structures, as illustrated in Fig.~\ref{fig:intro}. 
Given a 3DGS representation reconstructed from an input scene, our method applies the target style through a decoupled optimization process that alternately refines geometry and color parameters.
The decoupled geometry and color optimization is driven by the proposed geometry-aware contrastive feature matching (GCFM). To ensure a holistic alignment, GCFM leverages joint representations from color, depth, and edge modalities. A contrastive objective then guides the optimization by pulling each rendered feature toward its most similar counterpart in the style image while pushing it away from the most dissimilar one.
Experimental results demonstrate that our method achieves superior performance both qualitatively and quantitatively.
Our main contributions are summarized as follows.
\begin{itemize}

    \item[$\bullet$] We propose a simple yet effective geometry-aware style transfer framework for 3DGS, built upon a decoupled optimization scheme that alternates between color and geometry updates for stable and consistent stylization.
    

    \item[$\bullet$] We introduce GCFM, a geometry-aware contrastive feature matching scheme that aligns the scene’s geometry with the style’s distinctive structures by optimizing a contrastive loss across multi-modal representations incorporating color, depth, and edge cues.

    \item[$\bullet$] Experimental results demonstrate that our method produces visually faithful stylization and consistently outperforms existing approaches in both qualitative and quantitative evaluations.
\end{itemize}

\section{Related Work}
\label{sec:related}
\subsection{Neural Style Transfer}
\label{ssec:imageStyle}
Gatys~\etal~\cite{Gatys} pioneered neural style transfer by using Gram matrices of CNN feature activations to encode image style, capturing correlations between feature channels while discarding spatial information.
Following this work, research expanded rapidly in several directions: feed-forward networks for real-time style transfer~\cite{Johnson,Ulyanov}, adaptive instance normalization (AdaIN)~\cite{AdaIN} for arbitrary style transfer, and GAN-based approaches for unpaired domain-to-domain translation~\cite{CycleGAN,MUNIT,DRIT}. Recently, diffusion models enable text-driven and highly controllable style transfer with unprecedented fidelity~\cite{StyleDiffusion}. Despite these advances, image-based methods remain inherently limited to single-view domains, making them unsuitable for 3D scenarios where strict multi-view consistency is essential.

\subsection{3D Style Transfer}
\label{ssec:3DStyle}
3D style transfer aims to achieve consistent stylization across novel views. 
Early studies explored explicit scene representations such as point clouds~\cite{pointCloud} and meshes \cite{styleMesh}. 
With the emergence of NeRFs~\cite{NeRF}, subsequent approaches embedded style information into volumetric representations to achieve multi-view consistency~\cite{ARF,StylizedNeRF,StyleRF,CoARF,Geometry}. 
However, the high-dimensional implicit nature and the slow rendering speed of NeRF posed challenges for real-time applications and direct geometric manipulation.

The advent of 3DGS offers a compelling alternative with real-time rendering capabilities and explicit geometric primitives. In~\cite{StyleGaussian}, zero-shot style transfer was introduced, where VGG features are embedded into Gaussian primitives to maintain view coherence. Based on~\cite{StyleGaussian}, Kov{\'a}cs~\etal~\cite{GStyle} pruned and split Gaussians and incorporated multi-scale style losses for enhanced style fidelity.
In~\cite{StylizedGS}, user-controllable parameters for stylization (scale, region, and color) and losses to avoid artifacts were introduced. Liu~\etal~\cite{ABCGS} improved global style consistency by aligning scene content with style features using segmentation masks and regularization. Galerne~\etal~\cite{SGSST} optimized only the constant color component of each Gaussian to match multi-scale style statistics. Recently, Howil~\etal~\cite{clipgaussian} used CLIP guidance to enable both image- and text-guided style transfer across images, videos, 3D objects, and even 4D dynamic scenes within a unified Gaussian splatting framework.

Despite recent progress, most 3DGS-based style transfer methods remain limited to appearance adaptation. Modifying geometry is highly unstable and can easily distort the underlying scene structure, so most methods adopt a conservative design that preserves geometry. For example, \cite{StyleGaussian} and \cite{SGSST} update only color attributes while keeping geometry fixed; \cite{GStyle} introduces limited geometric adaptation through Gaussian splitting during pre-processing and fine-tuning stages; and \cite{StylizedGS} and \cite{ABCGS} employ depth preservation losses to maintain geometric fidelity rather than to actively incorporate geometry as a stylistic element. \cite{clipgaussian} enables joint optimization of color and geometry, but it still places primary emphasis on appearance adaptation, leaving geometry-oriented stylization relatively underexplored. While these strategies yield stable results, they fall short of treating geometry as a medium of style. 
In this work, we address this gap by exploring geometric stylization as a means to achieve more expressive and structurally consistent 3D style transfer.

\begin{figure*}[t] 
\centering 
\includegraphics[width=1.0\linewidth]{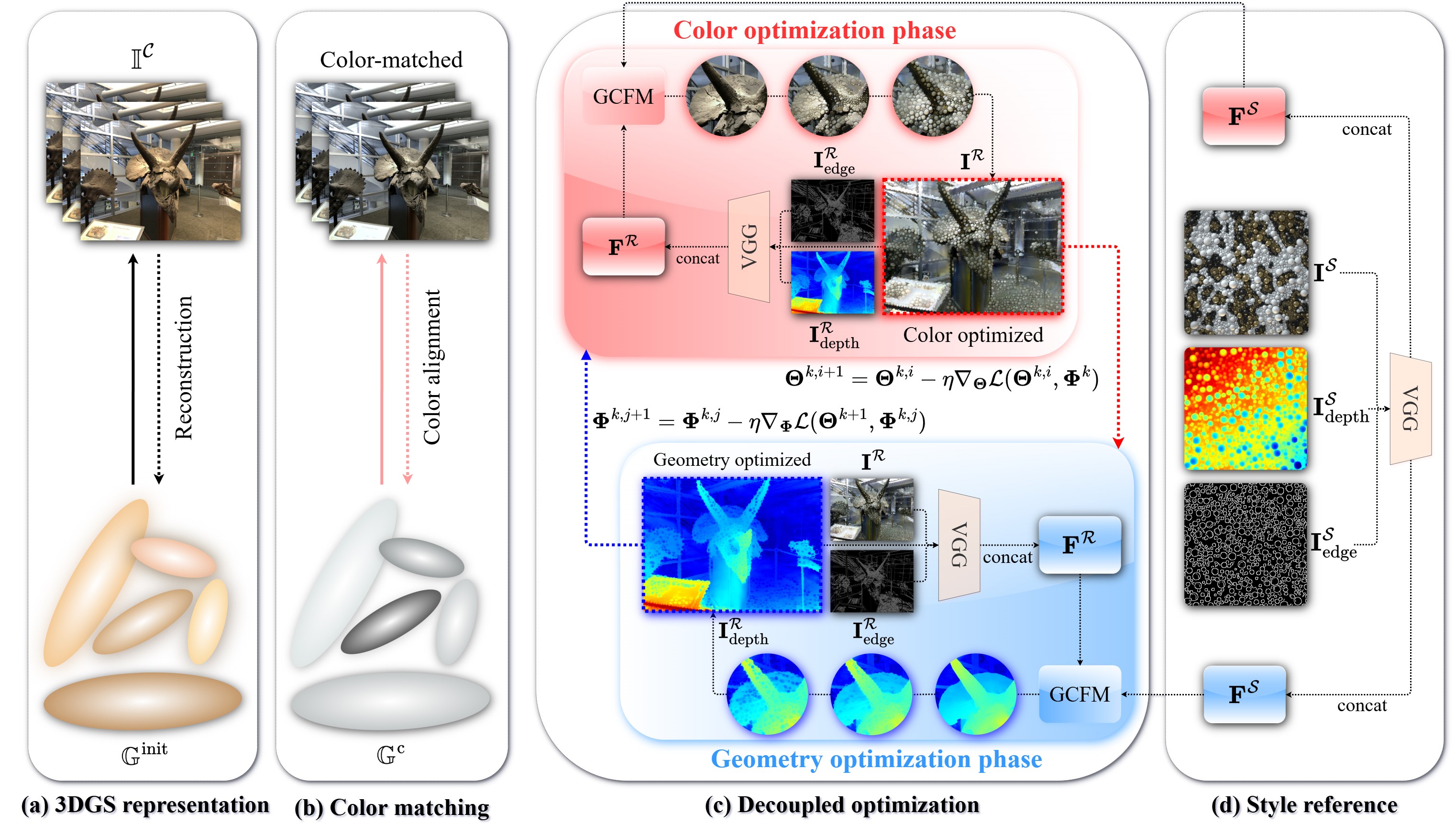} 
\vspace{-7mm}
\caption{Overview of the proposed geometry-aware style transfer framework for 3DGS. The pipeline consists of (a) 3DGS representation, (b) color matching, and (c) decoupled optimization between color and geometry optimization phases guided by multi-modal (color, depth, edge) features from (d) the style reference via the proposed GCFM.}
\label{fig:framework}
\vspace{-6mm}
\end{figure*}

\vspace{-2mm}
\section{Proposed Method}
\label{sec:proposed}
\begin{figure*}[t] 
\centering 
\includegraphics[width=0.92\linewidth]{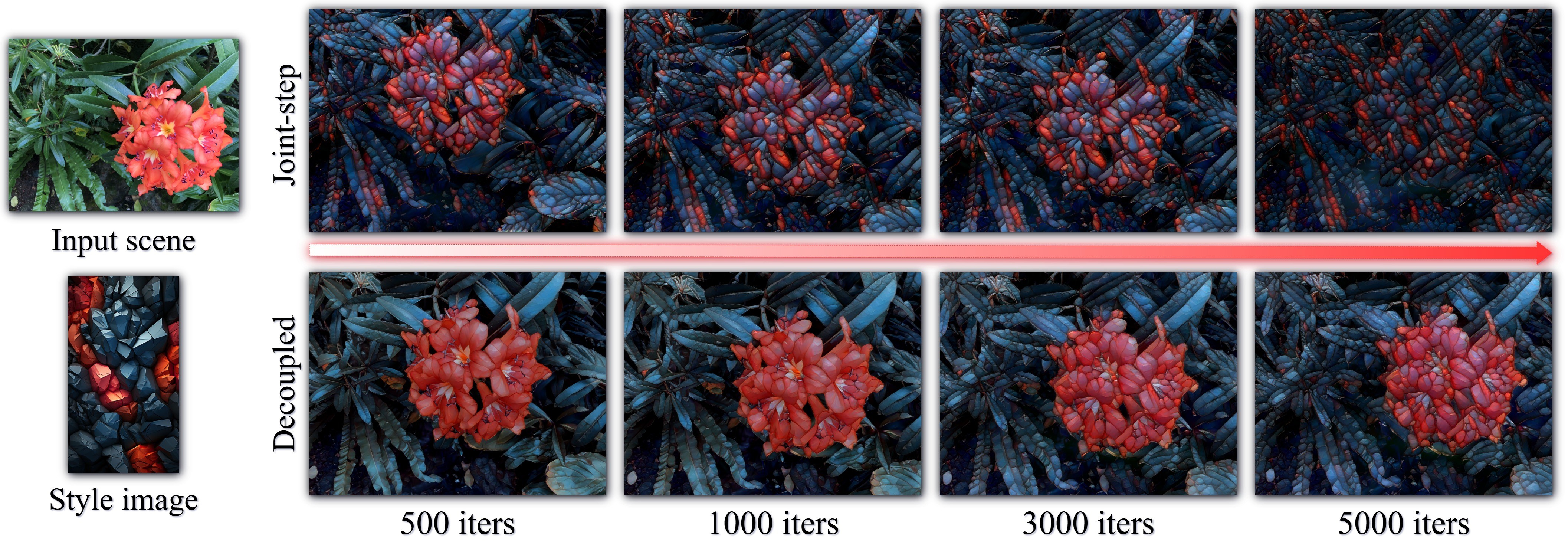}
\vspace{-4mm}
\caption{Comparison between joint-step (top) and decoupled (bottom) optimization.
The joint-step optimization updates color and geometry simultaneously, while our decoupled optimization alternates them (both run for the same total iterations).
}
\vspace{-9mm}
\label{fig:bilevel}
\end{figure*}
The goal of this paper is to transfer the artistic style from a 2D style image $\mathbf{I}^{\mathcal{S}}$ to a 3D scene constructed from a set of content images $\mathbb{I}^{\mathcal{C}} = \{ \mathbf{I}_i^{\mathcal{C}} \}_{i=1}^M$ with spatial and geometric coherence.
Fig.~\ref{fig:framework} illustrates the overall framework of the proposed style transfer method leveraging 3DGS~\cite{3DGS}. 
As a preliminary step,
a 3DGS representation $\mathbb{G}^{\textrm{init}} = \{ \mathbf{G}_n \coloneq (\mathbf{\theta}_n,\mathbf{\mu}_n, \mathbf{\Sigma}_n, \alpha_n ) \}_{n=1}^N$ is reconstructed from $\mathbb{I}^{\mathcal{C}}$, where each Gaussian primitive $\mathbf{G}_n$ is defined by the color representation $\mathbf{\theta}_n$, its 3D center position $\mathbf{\mu}_n \in \mathbb{R}^3$, its 3D covariance matrix $\mathbf{\Sigma}_n \in \mathbb{R}^{3 \times 3}$, and its opacity $\alpha_n \in [0,1]$.
We initialize the 3D style transfer by color matching~\cite{ARF}, obtaining $\mathbb{G}^{\mathrm{c}}$. This process aligns the color statistics of the content images in $\mathbb{I}^{\mathcal{C}}$ with those of the style image $\mathbf{I}^{\mathcal{S}}$ using a linear transformation, ensuring the Gaussian primitives start with a color distribution closer to the target style.
Then, we transfer the target style using a decoupled optimization scheme, which alternatively refines the color and geometric parameters. This alternating process is key to stabilizing the transfer process in 3DGS.
Both optimization phases are driven by the proposed GCFM, which effectively guides both the color and geometry refinement.
In the following sections, we describe our method in detail.

\subsection{Decoupled Optimization for 3D Style Transfer}
\label{ssec:bi-level}
Most existing 3DGS-based style transfer methods~\cite{ABCGS,GStyle,SGSST,StyleGaussian,StylizedGS} concentrate on transferring appearance attributes, whereas geometric stylization has remained a challenging problem. Let the Gaussian representation $\mathbb{G}$ decomposed into color parameters $\mathbf{\Theta} = \{ \mathbf{\theta}_n \}_{n=1}^{N}$ and geometry parameters $\mathbf{\Phi} = \{ \mathbf{\phi}_n \}_{n=1}^{N}$. Our ultimate goal is to jointly optimize the color parameters $\mathbf{\Theta}$ and the geometry parameters $\mathbf{\Phi}$ to minimize a global style transfer loss $\mathcal{L}(\mathbf{\Theta}, \mathbf{\Phi})$.
For the color parameters, we optimize only the zeroth-order (DC) component of the spherical harmonics color representation, which corresponds to the base RGB color, while keeping the higher-order coefficients fixed, following \cite{SGSST}. The geometry parameters are defined as $\mathbf{\phi}_n = \{\mathbf{\mu}_n,\mathbf{\Sigma}_n,\alpha_n\}$ for position, covariance, and opacity.

However, a naive simultaneous optimization of both parameter sets often results in interference between the color and geometry updates. This leads to color over-stylization, in which the optimization process sacrifices the essential structure of the scene due to a misguided attempt to embed low-level stylistic patterns directly into the geometry.
As illustrated in the first row of Fig.~\ref{fig:bilevel}, the flower's structure is progressively corrupted as iterations proceed, with exaggerated color patterns ignoring the underlying geometry and ultimately causing a complete loss of perceptual fidelity. 

To overcome these issues and enable stable geometry-aware style transfer, we introduce a \emph{decoupled optimization scheme}. The entire optimization proceeds for $K$ outer cycles. Within each cycle $k$, we break the global problem into two alternating subproblems as follows:
\begin{align}
\mathbf{\Theta}^{k+1} &= \arg\min_{\mathbf{\Theta}} \mathcal{L}(\mathbf{\Theta},\mathbf{\Phi}^k), \label{eq:step1} \\
\mathbf{\Phi}^{k+1} &= \arg\min_{\mathbf{\Phi}} \mathcal{L}(\mathbf{\Theta}^{k+1},\mathbf{\Phi}).   \label{eq:step2}
\end{align}
In this two-phase process, we first update the color parameters $\mathbf{\Theta}$ using the geometry fixed from the previous cycle, $\mathbf{\Phi}^k$.
Subsequently, we perform geometric stylization by updating the geometry parameters $\mathbf{\Phi}$ using the newly refined colors, $\mathbf{\Theta}^{k+1}$.
In practice, we approximate the solutions to these subproblems in Eqs.~(\ref{eq:step1}) and (\ref{eq:step2}) by performing a fixed number of inner gradient descent steps for each phase. Each outer cycle consists of two sequential phases:
\paragraph{Color optimization phase.}
To solve for $\mathbf{\Theta}^{k+1}$, we fix the geometry parameters $\mathbf{\Phi}^k$ and update the color parameters for $N_\mathrm{c}$ inner steps. For $i=0,\dots,N_\mathrm{c}-1$, the color parameters are updated as
\begin{equation}
\mathbf{\Theta}^{k,i+1}
= \mathbf{\Theta}^{k,i} - \eta \nabla_{\mathbf{\Theta}}
\mathcal{L}(\mathbf{\Theta}^{k,i}, \mathbf{\Phi}^k), \label{eq:color_update}
\end{equation}
where $\eta$ is the learning rate. The final color parameters after $N_\mathrm{c}$ steps, denoted as $\mathbf{\Theta}^{k+1} = \mathbf{\Theta}^{k,N_\mathrm{c}
}$, are then used in the following phase.
\paragraph{Geometry optimization phase.}
Subsequently, to solve for $\mathbf{\Phi}^{k+1}$, we hold the newly updated color parameters $\mathbf{\Theta}^{k+1}$ fixed and refine the geometry parameters for $N_\mathrm{g}$ inner steps. For $j=0,\dots,N_\mathrm{g}-1$, the geometry parameters are updated as
\begin{equation}
\mathbf{\Phi}^{k,j+1}
= \mathbf{\Phi}^{k,j} - \eta \nabla_{\mathbf{\Phi}}
\mathcal{L}(\mathbf{\Theta}^{k+1}, \mathbf{\Phi}^{k,j}). \label{eq:geo_update}
\end{equation}
The resulting geometry parameters, $\mathbf{\Phi}^{k+1} = \mathbf{\Phi}^{k,N_\mathrm{g}}$ are then carried forward to the next outer cycle, completing one full round of optimization.

By alternating color and geometry updates in this way, the scheme mitigates optimization conflicts between the two optimization processes and establishes a mutual guidance loop: color updates provide a stable reference for geometry refinement, and refined geometry supports consistent color adaptation. As a result, the optimization converges more stably and achieves reliable geometry-aware style transfer in 3DGS, as illustrated in the second row of Fig.~\ref{fig:bilevel}.

\subsection{Geometry-Aware Contrastive Feature Matching (GCFM)}

\begin{figure}[t]
\centering
\includegraphics[width=\columnwidth]{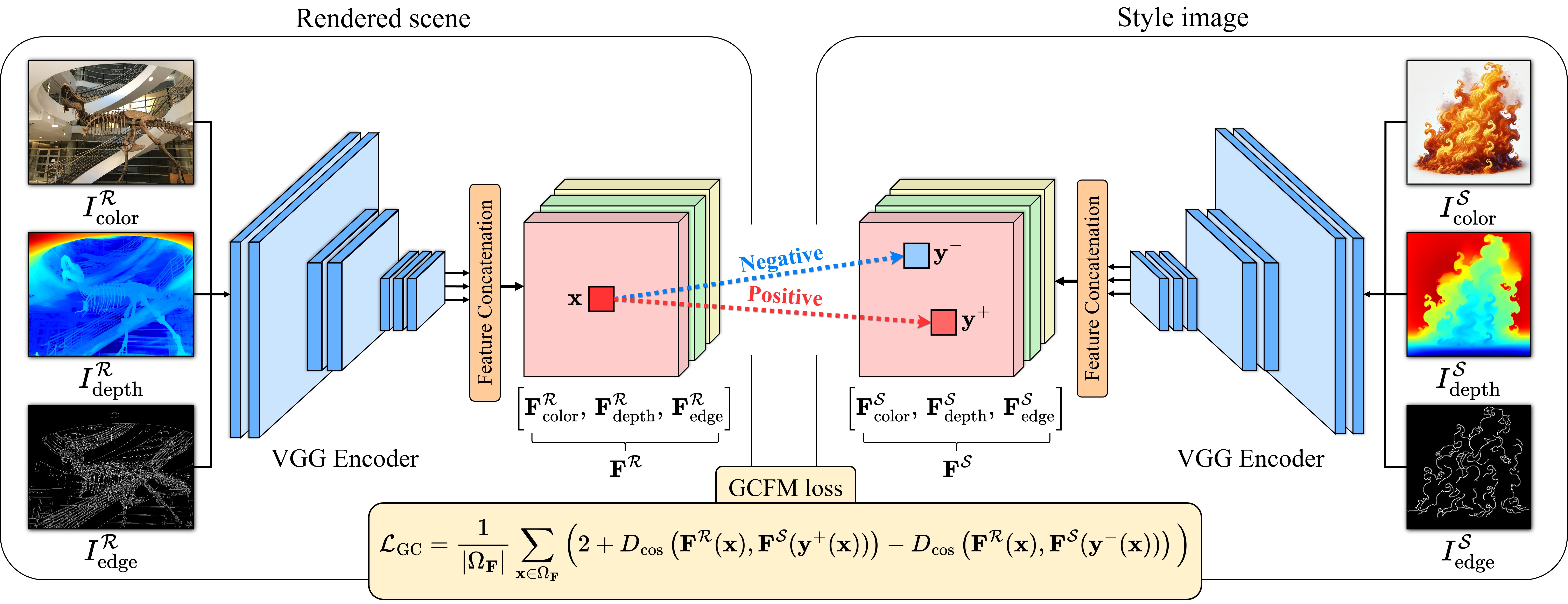}
\vspace{-7mm}
\caption{Conceptual illustration of the GCFM process.
}
\label{fig:Contrast NNFM}
\vspace{-0.7cm}
\end{figure}

Feature matching-based style transfer~\cite{ARF} has been widely adopted in many 3D style transfer approaches~\cite{ABCGS,Geometry,CoARF}. 
In this work, we utilize a multi-modal representation for feature matching, incorporating color, depth, and edge cues to enhance geometry awareness.
Depth maps can provide coarse spatial information, but often suffer from noise and missing details. Edge maps, on the other hand, highlight distinct structural boundaries and facilitate the alignment of shapes at the boundary level between content and style images. By jointly leveraging color, depth, and edge modalities, the proposed GCFM offers complementary guidance that enforces consistent 3D style transfer and alleviates the limitations of each individual cue.
Specifically, the depth map of the rendered image is obtained directly from the 3DGS rendering process, while the depth map of the style image is estimated using \cite{depthAnythingV2}.
Since complex textures and fine patterns, which are common in style images, tend to produce noisy edges in the color domain, we extract more structurally reliable edges from the depth map of the style image. The final edge maps are thus generated using the standard Canny detector~\cite{canny} from two different sources: the rendered image and the depth map of the style image.

Fig.~\ref{fig:Contrast NNFM} presents an overview of the GCFM process. For a given source image $\mathbf{I}^s$, where $s \in \{\mathcal{R}, \mathcal{S}\}$ denotes the rendered and style images, respectively, we use a pretrained VGG network~\cite{VGG} to extract deep features from the three modalities: color ($\mathbf{F}_{\mathrm{color}}^{s}$), depth ($\mathbf{F}_{\mathrm{depth}}^{s}$), and edge ($\mathbf{F}_{\mathrm{edge}}^{s}$).
Matching these features independently, however, can lead to inconsistencies, where appearance, shape, and edge patterns fail to align coherently. 
Therefore, to ensure holistic alignment, we employ a joint feature matching strategy by concatenating the features from all three modalities.
For a given source $s$, a joint feature map is defined as $\mathbf{F}^{s} = \big[ \mathbf{F}_{\mathrm{color}}^{s},\, \mathbf{F}_{\mathrm{depth}}^{s},\, \mathbf{F}_{\mathrm{edge}}^{s} \big]$, where $[\cdot,\cdot,\cdot]$ indicates channel-wise concatenation.

With the joint feature maps defined, it is required to align the structural features between the rendered and style images. 
Existing methods define positive pairs, \ie, structurally similar features that are pulled closer in the embedding space~\cite{ARF,Geometry}. However, we argue that for effective 3D style transfer, it is insufficient to only reflect such similarities.
To capture more discriminative style representations, the learning process should also explicitly contrast the rendered features against dissimilar features from the style source. Therefore, we additionally define negative pairs, representing dissimilar features that should be pushed apart, enabling the learning process to capture more discriminative style representations.
We identify these pairs for each spatial location $\mathbf{x}=(u,v)$ in the rendered feature map $\mathbf{F}^\mathcal{R}$. The positive sample is its nearest neighbor in the style feature map $\mathbf{F}^\mathcal{S}$ while the negative sample is the farthest one. We find the spatial location of these samples, $\mathbf{y}^{+}(\mathbf{x})$ and $\mathbf{y}^{-}(\mathbf{x})$, by respectively minimizing and maximizing a feature distance as follows:
\begin{align}
\mathbf{y}^{+}(\mathbf{x}) = \underset{\mathbf{y}}{\arg\min} \;
D_{\cos}\big(\mathbf{F}^{\mathcal{R}}(\mathbf{x}), \mathbf{F}^{\mathcal{S}}(\mathbf{y})\big),\\
\mathbf{y}^{-}(\mathbf{x}) = \underset{\mathbf{y}}{\arg\max} \;
D_{\cos}\big(\mathbf{F}^{\mathcal{R}}(\mathbf{x}), \mathbf{F}^{\mathcal{S}}(\mathbf{y})\big),
\end{align}
where $D_{\cos} (\cdot,\cdot) = 1 - \cos(\cdot,\cdot)$ represents the cosine distance. 
Based on the matched positive and negative indices, we formulate a single contrastive objective, called GCFM loss. The core principle is to treat each feature $\mathbf{F}^\mathcal{R}(\mathbf{x})$ from the rendered image as an anchor, pulling it toward its positive style counterpart $\mathbf{F}^\mathcal{S}(\mathbf{y}^{+}(\mathbf{x}))$ while pushing it away from its negative counterpart $\mathbf{F}^\mathcal{S}(\mathbf{y}^{-}(\mathbf{x}))$ during optimization. The final GCFM loss is thus defined as follows:
\begin{align}
\mathcal{L}_{\mathrm{GC}} = \frac{1}{\vert \mathrm{\Omega}_{\mathbf{F}} \vert} \sum_{\mathbf{x} \in \mathrm{\Omega}_{\mathbf{F}}} &\Big( 2 + D_{\cos}\left( \mathbf{F}^\mathcal{R}(\mathbf{x}), \mathbf{F}^\mathcal{S}(\mathbf{y}^{+}(\mathbf{x})) \right) \nonumber\\ & - D_{\cos}\left( \mathbf{F}^\mathcal{R}(\mathbf{x}), \mathbf{F}^\mathcal{S}(\mathbf{y}^{-}(\mathbf{x})) \right) \Big),
\end{align}
where $\mathrm{\Omega}_\mathbf{F}$ is a set of all spatial locations of the feature map.
\vspace{-0.3cm}

\subsection{Loss Function}
\label{ssec:loss}
In this section, we present the loss function $\mathcal{L}(\mathbf{\Theta}, \mathbf{\Phi})$ used for the proposed method to optimize the color parameters $\mathbf{\Theta}$ and the geometry parameters $\mathbf{\Phi}$. For the sake of brevity, we simply denote the loss function as $\mathcal{L}$, which is given by
\begin{equation}
\mathcal{L}
= \lambda_{\mathrm{GC}}
\mathcal{L}_{\text{GC}}
+ \lambda_{\mathrm{cont}} \mathcal{L}_{\mathrm{cont}}
+ \lambda_{\mathrm{TV}} \mathcal{L}_{\mathrm{TV}}
+ \delta_{\text{reg}} \, \mathcal{L}_{\text{reg}},
\end{equation}
where $\lambda_{\mathrm{GC}}$, $\lambda_{\mathrm{cont}}$, and $\lambda_{\mathrm{TV}}$ are empirically chosen hyper-parameters that control the relative contribution of each loss term. Here, $\delta_{\text{reg}}=1$ during the geometry optimization phase, and $\delta_{\text{reg}}=0$ during the color optimization phase. $\mathcal{L}_{\mathrm{cont}}$, $\mathcal{L}_{\mathrm{TV}}$, and $\mathcal{L}_{\text{reg}}$ denote the content preservation loss, total variation (TV) loss, and geometry regularization loss, respectively. Each term is detailed as follows:
\paragraph{Content preservation loss.} To ensure that the resulting stylized image preserves the structural fidelity of the original content, we adopt a perceptual loss based on VGG feature embeddings~\cite{VGG}. To this end, we minimize the distance between the VGG features of the rendered image $\mathbf{I}^{\mathcal{R}}$ and those of the content image $\mathbf{I}^{\mathcal{C}}$ as follows:
\begin{equation}
\mathcal{L}_{\mathrm{cont}} = \|\psi(\mathbf{I}^{\mathcal{R}}) - \psi(\mathbf{I}^{\mathcal{C}})\|^2_2,
\end{equation}
where $\psi(\cdot)$ denotes the feature representation extracted from a pre-trained VGG-16 network.
The content preservation loss maintains the original structure of the underlying scene while allowing visual adaptation to the style image.

\paragraph{TV loss.} To further improve the visual quality of the stylized results, we impose a TV regularization that encourages local spatial smoothness in the rendered image. Minimizing the TV loss effectively suppresses high-frequency noise and reduces artifacts along object boundaries. The TV loss is given by

\begin{equation}
\mathcal{L}_{\mathrm{TV}} = \frac{1}{\vert\mathrm{\Omega}_{\mathbf{I}}\vert} \sum_{\mathbf{x} \in \mathrm{\Omega}_{\mathbf{I}}}  \big[\|\mathbf{I}^{\mathcal{R}}(\mathbf{x}) - \mathbf{I}^{\mathcal{R}}(\mathbf{x}+\mathbf{e}_\mathrm{h})\|^2 + \|\mathbf{I}^{\mathcal{R}}(\mathbf{x}) - \mathbf{I}^{\mathcal{R}}(\mathbf{x}+\mathbf{e}_\mathrm{v})\|^2\big],
\end{equation}
where $\mathrm{\Omega}_\mathbf{I}$ is a set of all spatial locations of the rendered image, and $\mathbf{e}_\mathrm{h}=(0,1)$ and $\mathbf{e}_\mathrm{v}=(1,0)$ are the unit offset vectors for the horizontal and vertical neighbors, respectively.

\paragraph{Geometry regularization loss.}
Unconstrained geometry optimization may cause Gaussian primitives to degenerate into undesirable configurations, 
such as excessive expansion that produces global blurring or collapse into elongated shapes. To prevent such artifacts, we adopt a geometry regularization loss that penalizes deviations in opacity, scale, and rotation from their initial color-matched states, and additionally enforces depth consistency:
\begin{equation}
\mathcal{L}_{\mathrm{reg}} 
= \lambda_{\alpha}\,\|\mathbf{\alpha} - \mathbf{\alpha}^{\mathrm{c}}\|_{2}
+ \lambda_{\mathrm{s}}\,\|\mathbf{s} - \mathbf{s}^{\mathrm{c}}\|_{2} + \lambda_{\mathrm{r}}\,\|\mathbf{r} - \mathbf{r}^{\mathrm{c}}\|_{2}
+ \lambda_{\mathrm{d}}\,\|\mathbf{D}^{\mathcal{R}} - \mathbf{D}^{\mathrm{c}}\|,
\end{equation}
where the weighting coefficients $\lambda_{\alpha}$, $\lambda_{\mathrm{s}}$, $\lambda_{\mathrm{r}}$, and $\lambda_{\mathrm{d}}$ are chosen empirically. 
Here, $\mathbf{\alpha}$, $\mathbf{s}$, and $\mathbf{r}$ denote the opacity, scale, and rotation parameters of all Gaussian primitives, respectively, where the superscript ${\mathrm{c}}$ indicates their values after color matching with $\mathbb{G}^{\mathrm{c}}$. $\mathbf{D}^{\mathcal{R}}$ denotes the rendered depth map from the current 3DGS model, and $\mathbf{D}^{\mathrm{c}}$ represents the reference depth map obtained from $\mathbb{G}^{\mathrm{c}}$, respectively. Overall, this regularization anchors the stylization process to the original scene structure, enabling effective style-driven adaptations while preventing instability in geometry.

\begin{figure*}[t] 
\centering 
\includegraphics[width=0.92\linewidth]{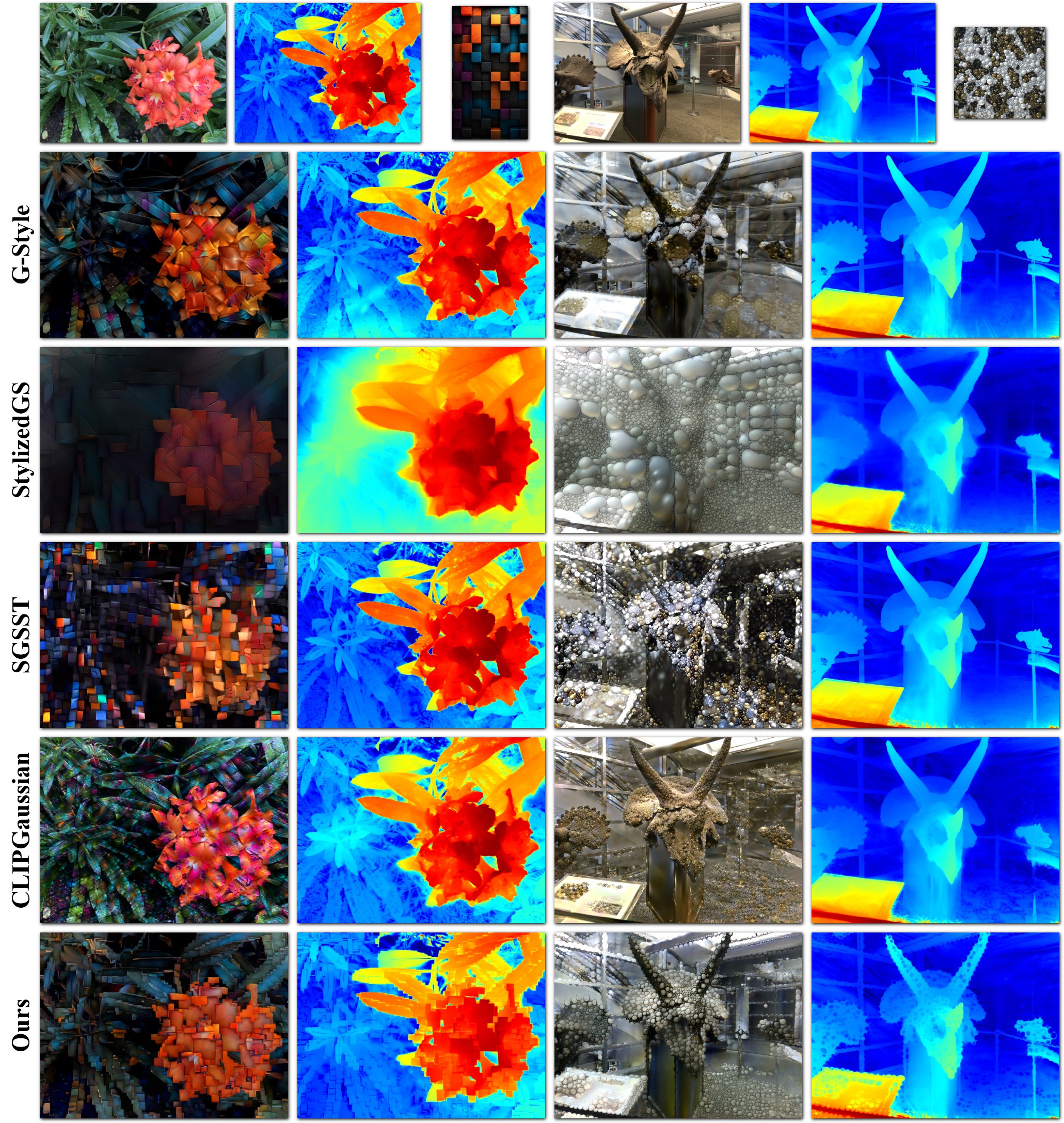} 
\vspace{-0.3cm}
\caption{Qualitative comparison of the \textit{Flower} and \textit{Horns} scenes. The top row shows the input scenes, rendered depth maps, and the corresponding style image for each scene. The subsequent rows present stylized results produced by different methods, including G-Style~\cite{GStyle}, StylizedGS~\cite{StylizedGS}, SGSST~\cite{SGSST}, CLIPGaussian~\cite{clipgaussian}, and Ours. 
}
\vspace{-0.7cm}
\label{fig:compare_results2} 
\end{figure*}

\section{Experiments}
\label{sec:exp}
\subsection{Datasets}
\label{ssec:dataset}
We evaluate our 3DGS-based style transfer method across a total of 72 scene–style combinations (8 scenes × 9 styles).

\paragraph{Scenes.}
We use three representative multi-view scene datasets: LLFF~\cite{llff} (4 forward-facing scenes), Tanks \& Temples (T\&T)~\cite{tnt} (2 complex indoor/outdoor 360° scenes), and MipNeRF-360~\cite{mipnerf360} (2 large-scale unbounded scenes).
Together, these eight scenes provide a wide range of geometric structures and camera trajectories, from small bounded scenes to large-scale 360° environments, allowing robust evaluation of geometry-aware stylization.
\vspace{-0.3cm}
\paragraph{Styles.}
We use a set of nine style exemplars gathered from multiple sources—four from \cite{Geometry}, three from WikiArt~\cite{wikiart}, one from \cite{clipgaussian}, and one curated from publicly available online art collections~\cite{freepik_waterwaves}. To ensure stylistic diversity and generalization, these exemplars are deliberately drawn from heterogeneous sources, covering both classical and modern art movements as well as 3D-aware styles. The selection emphasizes spatially structured patterns to evaluate the model’s ability to capture geometric as well as appearance-based stylistic characteristics.
\vspace{-0.4cm}


\subsection{Implementation Details}
\label{sec:implementation}

Our implementation is built upon the public PyTorch~\cite{pytorch} codebase of 3DGS~\cite{3DGS}.
We adopt the proposed decoupled optimization scheme for geometry-aware style transfer, where color and geometry parameters are alternately updated. Algorithm~\ref{alg:pseudo} in the supplementary material summarizes the overall optimization procedure, illustrating how the color and geometry parameters are updated in an alternating manner under the proposed framework.
Unless otherwise specified, the loss weights are set as $\lambda_{\mathrm{GC}} = 2.0$, 
$\lambda_{\mathrm{cont}} = 1\times10^{-3}$, 
$\lambda_{\mathrm{TV}} = 0.02$, 
$\lambda_{\alpha} = 1.0$, 
$\lambda_{\mathrm{s}} = 1.0$
$\lambda_{\mathrm{r}} = 1.0$, and
$\lambda_{\mathrm{d}} = 10.0$.
We perform the stylization process for $K = 30$ outer cycles using the Adam optimizer~\cite{adam}, 
with $N_{\mathrm{c}} = 10$ color updates and $N_{\mathrm{g}} = 90$ geometry updates per cycle, totaling $3{,}000$ iterations. We set the learning rate $\eta = 1\times10^{-4}$ for all experiments, except for the T\&T dataset~\cite{tnt}, where it is reduced to $2.5\times10^{-5}$ for stable convergence, following \cite{GStyle} which adopted a similar adjustment for scenes with higher geometric complexity. All experiments are conducted on a single NVIDIA RTX A6000 GPU with 48 GB of memory.

\vspace{-0.3cm}

\subsection{Comparison with State-of-the-Arts}
\label{ssec:comparison}

\begin{table}[t]
\centering
\caption{
Quantitative comparison of style transfer quality (SIFID) and multi-view consistency. \textbf{Bold} and \underline{underline} indicate the best and the second-best results.
}
\vspace{-0.2cm}
\setlength\tabcolsep{6.0pt}
\resizebox{0.62\columnwidth}{!}{
\begin{tabular}{cccccc}
    \toprule
    \multirow{2}{*}{Methods} & \multirow{2}{*}{SIFID↓} & \multicolumn{2}{c}{\makecell{Short-Range \\ Consistency}} & \multicolumn{2}{c}{\makecell{Long-Range \\ Consistency}} \\
    \cmidrule(lr){3-4} \cmidrule(lr){5-6}
            & & LPIPS↓ & RMSE↓ & LPIPS↓ & RMSE↓ \\
    \midrule
    StyleGaussian~\cite{StyleGaussian} & 3.3405 & 0.050 & 0.053 & 0.141 & 0.114 \\
    G-Style~\cite{GStyle}             & \underline{1.2955} & 0.045 & 0.035 & 0.118 & 0.085 \\
    StylizedGS~\cite{StylizedGS}      & 2.2439 & \textbf{0.028} & \textbf{0.021} & \textbf{0.072} & \textbf{0.062} \\
    SGSST~\cite{SGSST}                & 1.3936 & 0.048 & 0.048 & 0.129 & 0.109 \\
    CLIPGaussian~\cite{clipgaussian}  & 3.3685 & 0.048 & 0.046 & 0.128 & 0.108 \\
    \midrule
    \textbf{Ours}                     & \textbf{1.1736} & \underline{0.042} & \underline{0.034} & \underline{0.114} & \underline{0.084} \\
    \bottomrule
\end{tabular}
}
\vspace{-0.5cm}
\label{tab:combined_quantitative}
\end{table}

\paragraph{Qualitative comparison.} 
Fig.~\ref{fig:compare_results2} compares our method with state-of-the-art 3DGS-based style transfer approaches~\cite{SGSST,GStyle,StylizedGS,clipgaussian}. Unlike previous methods that primarily focus on color-based appearance transfer, our approach jointly adapts both appearance and geometric structure, producing visually richer and structurally coherent 3D stylization across all evaluated scenes. For instance, in the \textit{Flower} scene on the left, our method successfully transforms the object’s shape to reflect the block-like structural patterns of the target style, rather than merely applying color changes. In the \textit{Horns} scene on the right, our method transforms the original surfaces into particle-like structures while preserving fine geometric details and clear depth variations around the horn. This enhanced structural expressiveness leads to more immersive and visually compelling results overall.

\paragraph{Quantitative comparison.}
Table~\ref{tab:combined_quantitative} summarizes the quantitative evaluation using the Single Image Fréchet Inception Distance (SIFID)~\cite{SinGAN}, which measures the perceptual distance between each stylized image and its target style.
Unlike FID~\cite{FID}, which requires large image sets, SIFID is suitable for evaluating single-scene stylization results. Each result is rendered from multiple viewpoints and averaged to ensure robustness. A lower SIFID indicates closer alignment with the target style distribution. As shown in the table, our method achieves the lowest SIFID score (1.1736), demonstrating superior style fidelity and perceptual alignment with the target style compared to previous methods.

Also in Table~\ref{tab:combined_quantitative}, we further evaluate the multi-view consistency of different methods under both short- and long-range settings. Following~\cite{StyleGaussian}, one view is warped to another using optical flow~\cite{OF} with softmax splatting~\cite{splatting}, and stylization consistency is measured by masked RMSE and LPIPS~\cite{LPIPS}.
While StylizedGS~\cite{StylizedGS} reports the best consistency scores, this may stem from its overly smooth and desaturated renderings, which inherently favor consistency metrics rather than reflecting strong stylization effects.\footnote{A detailed comparison with StylizedGS, including qualitative analysis, is provided in Section \ref{supp:stylizedGS} of the supplementary material.} Excluding StylizedGS, our method achieves the best overall performance, ranking first in both LPIPS and RMSE across short- and long-range consistency settings. These results indicate that our framework achieves a better balance between stylization quality and multi-view consistency compared to other 3DGS-based methods.


\begin{figure*}[t]
\centering
\includegraphics[width=1.0\linewidth]{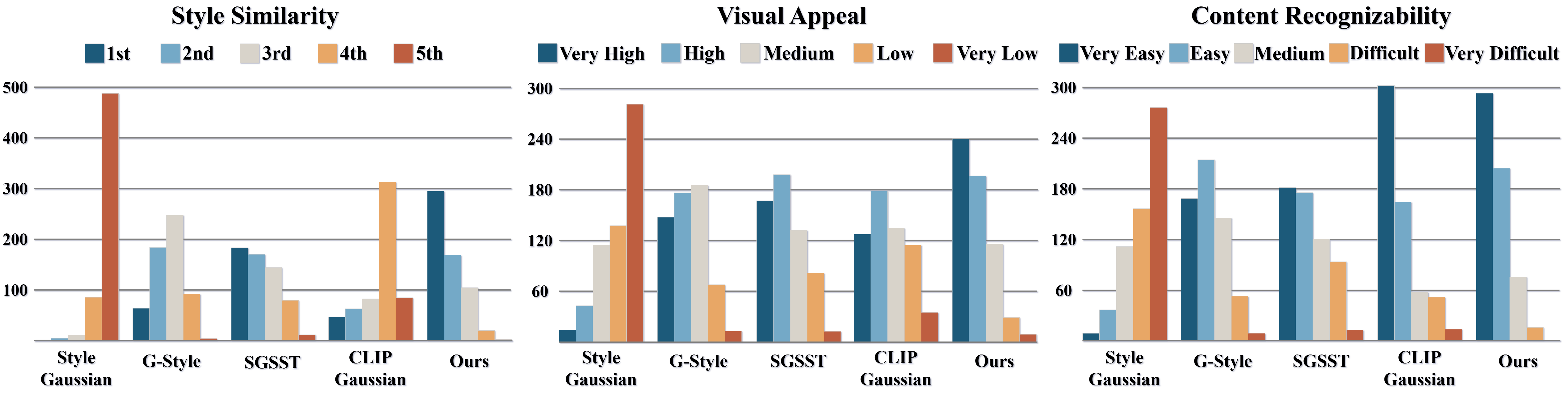}
\vspace{-0.2cm}
\caption{Results of the user study comparing StyleGaussian~\cite{StyleGaussian}, G-Style~\cite{GStyle}, SGSST~\cite{SGSST}, CLIPGaussian~\cite{clipgaussian}, and Ours.}
\label{fig:user_study_2}
\vspace{-0.6cm}
\end{figure*}

\subsection{User Study}
\label{ssec:userStudy}
To assess the perceptual quality of the stylized renderings, we conducted an online user study involving 37 participants. Our method was compared against four representative 3DGS-based style transfer methods~\cite{SGSST,GStyle,StyleGaussian,clipgaussian}. We used four distinct 3D scenes and four style reference images, each containing spatially consistent texture and geometric cues—such as blocky patterns and particle-like elements—two of which are shown in Fig.~\ref{fig:compare_results2}. For each result, we asked participants to rate their (1) style similarity to the reference style image, (2) overall visual appeal, and (3) content recognizability of the original scene.

The analyzed outcomes of the user study are shown in Fig.~\ref{fig:user_study_2}. Participants consistently rated our approach as the best across all three evaluation criteria, demonstrating its effectiveness in producing high-quality stylizations that are faithful to the target style while preserving scene content. 

\vspace{-0.4cm}

\begin{table}[t]
\centering
\caption{
    Comparison of average stylization times across various 3DGS-based methods.
    }
    \vspace{-0.2cm}
\resizebox{0.32\columnwidth}{!}{%
\begin{tabular}{ccc}
    \toprule
    Methods & Stylization Time \\
    \midrule
    StyleGaussian~\cite{StyleGaussian} & 
    410m 13s
    \\
    G-Style~\cite{GStyle} 
    &29m 05s
    \\
    StylizedGS~\cite{StylizedGS} & 
    \textbf{3m 32s}
    \\
    SGSST~\cite{SGSST} & 
    58m 52s\\
    CLIPGaussian~\cite{clipgaussian} & 
    22m 46s
    \\
    Ours & 
    \underline{14m 14s}
    \\
    \bottomrule
\end{tabular}
    }
    \label{tab:stylization_time}
    \vspace{-0.3cm}
\end{table}

\subsection{Complexity Analysis}
\label{sec:complexity}

Table~\ref{tab:stylization_time} compares the average stylization time of representative 3DGS-based methods.
While StyleGaussian~\cite{StyleGaussian} requires several hours of optimization for a single style, our method completes the stylization in about fourteen minutes, showing a substantial improvement in efficiency.
Although StylizedGS~\cite{StylizedGS} achieves the fastest runtime, it often compromises geometric consistency and style fidelity, as discussed in Section~\ref{supp:stylizedGS} of the supplementary material.
Overall, our approach maintains a favorable trade-off between computational speed and visual quality, providing a practical and scalable solution for 3D scene stylization. Detailed per-scene analysis on stylization time and resource usage is provided in Section~\ref{supp:complexity} of the supplementary material.

\vspace{-0.4cm}

\begin{figure}[t]
\centering
\includegraphics[width=0.65\columnwidth]{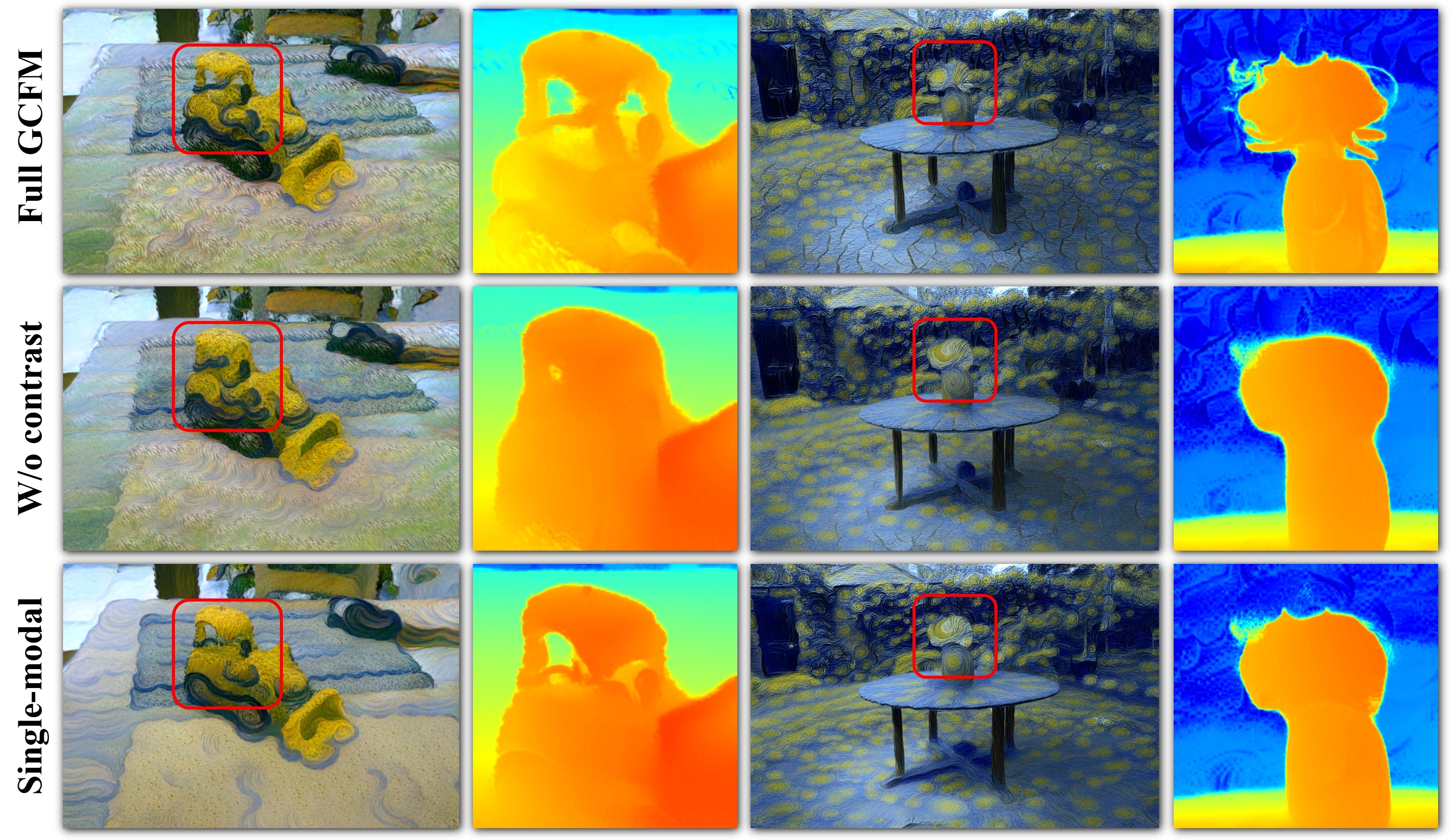}
\vspace{-0.2cm}
\caption{Ablation study on GCFM. The 1st and 3rd columns show the stylized rendered views, while the 2nd and 4th columns display the corresponding enlarged depth maps.
}
\label{fig:Ablation_GeoNNFM}
\vspace{-0.6cm}
\end{figure}

\subsection{Ablation Studies}
\label{ssec:ablation}
\paragraph{Ablation study on GCFM.}
Fig.~\ref{fig:Ablation_GeoNNFM} presents the ablation results of GCFM, showing how each component contributes to geometry-aware stylization. To capture geometric cues beyond appearance, GCFM integrates both depth and edge modalities into the feature representation. Additionally, a contrastive objective is introduced to enforce stronger feature alignment, enhancing the distinctiveness of geometry-related patterns. As shown in the figure, removing the contrastive objective (w/o contrast) weakens feature alignment, leading to a loss of structural distinctiveness in the depth maps. Similarly, using a single-modal representation (single-modal) results in oversimplified geometric structures, which indicates that multi-modal integration is essential for capturing fine geometric contours. Overall, the full GCFM yields sharper contours and faithfully reproduces the underlying geometric structures, confirming that both the contrastive loss and multi-modal feature integration are essential for capturing and utilizing complex scene geometry effectively.

\vspace{-0.3cm}

\begin{table}[t]
\centering
\caption{
    Ablation study on optimization strategies.
    }
\vspace{-0.3cm}  
\setlength\tabcolsep{6.0pt}
\resizebox{0.65\columnwidth}{!}{%
\begin{tabular}{ccccccc}
    \toprule
    \multirow{2}{*}{Methods} & \multirow{2}{*}{SIFID↓} & \multicolumn{2}{c}{\makecell{Short-Range \\ Consistency}} & \multicolumn{2}{c}{\makecell{Long-Range \\ Consistency}} \\
    \cmidrule(lr){3-4} \cmidrule(lr){5-6}
            & & LPIPS↓ & RMSE↓ & LPIPS↓ & RMSE↓ \\
    \midrule
    Only geometry & 2.5496 & 0.045 & 0.040 & \textbf{0.113} & 0.096 \\
    Only color & 1.5392 & \textbf{0.042} & 0.037 & 0.115 & \underline{0.086} \\
    Joint-step & \underline{1.4220} & \textbf{0.042} & \underline{0.036} & 0.122 & 0.087 \\
    Decoupled (Ours) & \textbf{1.1736} & \textbf{0.042} & \textbf{0.034} & \underline{0.114} & \textbf{0.084} \\
    \bottomrule
\end{tabular}
    }
    \label{tab:ablation_optimization}
    \vspace{-0.6cm}
\end{table}

\paragraph{Ablation study on optimization strategies.}
Table~\ref{tab:ablation_optimization} compares different optimization strategies for geometry-aware stylization.
When updating only geometry, the model primarily modifies structural parameters while keeping color parameters unchanged.
This results in poor stylization fidelity and the highest SIFID due to insufficient adaptation to the target style, and also causes noticeable cross-view inconsistency from unstable geometric updates.
In contrast, optimizing only color achieves lower SIFID by transferring color appearance more effectively, but it still fails to capture geometry-dependent style cues, leading to weak structural adaptation and reduced consistency across views.
The joint-step approach, which updates color and geometry simultaneously, achieves moderate improvement in SIFID but suffers from color over-stylization and unstable convergence, ultimately leading to inferior performance.
Finally, our decoupled optimization scheme achieves the best trade-off between stylization fidelity and geometric stability, showing the lowest SIFID and consistent improvements in both short- and long-range LPIPS and RMSE.
These results show that separating the color and geometry updates stabilizes optimization and enables harmonious integration of stylistic appearance and geometric adaptation.

Additional results and analyses are provided in Sections \ref{supp:user_study}–\ref{supp:additional_qualitative} of the supplementary material, including user study details, ablation studies, and extended qualitative results that further validate the effectiveness of the proposed method.
\vspace{-0.8cm}

\section{Conclusions}
In this work, we presented a geometry-aware style transfer framework for 3DGS that jointly stylizes both appearance and geometry, achieving stable optimization and consistent stylization results. To address the optimization instability caused by simultaneous updates of color and geometry, we introduced a decoupled optimization scheme that alternates between the two, enabling mutual guidance between appearance and structure. During this process, we employed the proposed GCFM loss, which integrates RGB, depth, and edge cues into a unified contrastive objective to effectively capture geometric and structural characteristics of the target style. Extensive experiments across various 3D scenes demonstrate that our method achieves superior performance in both stylization quality and multi-view consistency compared to recent 3DGS-based methods.

\section*{Acknowledgements}
This work was supported by the National Research Foundation of Korea(NRF) grant funded by the Korea Government(MSIT)(RS-2026-25473686, RS-2026-25489106).

\bibliographystyle{splncs04}
\bibliography{main}
\clearpage
\appendix 
\setcounter{page}{1}

\renewcommand\thesection{\Alph{section}}
\renewcommand\thetable{\Alph{table}}
\renewcommand\thefigure{\Alph{figure}}
\renewcommand\thealgorithm{\Alph{algorithm}}
\renewcommand\theequation{\Alph{equation}}

\setcounter{table}{0}
\setcounter{figure}{0}
\setcounter{equation}{0}

\begin{center}
    \Large \textbf{Geometry-Aware Style Transfer in 3D Gaussian Splatting \\}\par
    \Large \textbf{\textit{-- Supplementary Material --}}
\end{center}


\begin{algorithm}
  \caption{Geometry-aware style transfer in 3DGS}
  \begin{algorithmic}[1]
    \State \textbf{Input:} Content images $\mathbb{I}^{\mathcal{C}} = \{ \mathbf{I}_i^{\mathcal{C}} \}_{i=1}^M$, style image $\mathbf{I}^\mathcal{S}$
    \State \textbf{Output:} Stylized Gaussians $\mathbb{G}^{\mathrm{s}}=\{\mathbf{\Theta}^{\mathrm{s}},\mathbf{\Phi}^{\mathrm{s}}\}$
    \State $\mathbb{G}^{\mathrm{init}} \gets \textsc{Recon3DGS}(\mathbb{I}^{\mathrm{c}})$ 
    \State $\mathbb{G}^{\mathrm{c}}=\{\mathbf{\Theta}^0,\mathbf{\Phi}^0\} \gets \textsc{ColorMatching}(\mathbb{G}^{\mathrm{init}}, \mathbb{I}^{\mathcal{C}}, \mathbf{I}^{\mathcal{S}})$
    \For{$k=1,2,\ldots,K$} \Comment{outer cycle}
      \State $\mathbf{\Theta}\gets \mathbf{\Theta}^{k-1}$
      \State $\mathbf{\Phi}\gets \mathbf{\Phi}^{k-1}$
      \State // \textbf{Color optimization phase}
      \For{$i=1,2,\ldots,N_{\mathrm{c}}$}
        \State $v \gets \textsc{RandomView()}$
\State $\mathbf{I}^{\mathcal{R}} \gets \textsc{Render}(v,\{\mathbf{\Theta},\mathbf{\Phi}\})$ \Comment{to compute $\mathcal{L}$}
        \State $\mathcal{L} \gets \lambda_{\mathrm{GC}}\mathcal{L}_{\mathrm{GC}}+\lambda_{\mathrm{TV}}\mathcal{L}_{\mathrm{TV}}+\lambda_{\mathrm{cont}}\mathcal{L}_{\mathrm{cont}}$
        \State $\mathbf{\Theta} \gets \mathbf{\Theta}-\eta\,\nabla_{\mathbf{\Theta}}\mathcal{L}$ \Comment{$\mathbf{\Phi}$ is not updated}
      \EndFor
      \State $\mathbf{\Theta}^{k}\gets \mathbf{\Theta}$

      \State // \textbf{Geometry optimization phase}
      \For{$j=1,2,\ldots,N_{\mathrm{g}}$}
        \State $v \gets \textsc{RandomView()}$ 
\State $\mathbf{I}^{\mathcal{R}} \gets \textsc{Render}(v,\{\mathbf{\Theta},\mathbf{\Phi}\})$ \Comment{to compute $\mathcal{L}$}
        \State $\mathcal{L} \gets \lambda_{\mathrm{GC}}\mathcal{L}_{\mathrm{GC}}+\lambda_{\mathrm{TV}}\mathcal{L}_{\mathrm{TV}}+\lambda_{\mathrm{cont}}\mathcal{L}_{\mathrm{cont}}+\mathcal{L}_{\mathrm{reg}}$
        \State $\mathbf{\Phi} \gets \mathbf{\Phi}-\eta\,\nabla_{\mathbf{\Phi}}\mathcal{L}$ \Comment{$\mathbf{\Theta}$ is not updated}
      \EndFor
      \State $\mathbf{\Phi}^{k}\gets \mathbf{\Phi}$ 
    \EndFor
    \State \textbf{return} $\mathbb{G}^K=\{\mathbf{\Theta}^K,\mathbf{\Phi}^K\}$
  \end{algorithmic}
  \label{alg:pseudo} 
\end{algorithm}

\begin{table}[t]
\centering
\caption{
    Computational efficiency analysis. Summary of the average number of Gaussians, stylization time, and GPU memory usage for each scene.
    }
    \vspace{-0.3cm}
\resizebox{0.7\columnwidth}{!}{%
\begin{tabular}{ccccc}
    \toprule
    ~Scenes~ & ~\# of Gaussians (M)~ & ~Stylization Time~ & ~GPU Memory (GB)~ \\
    \midrule
    trex & 0.54 & 17m 54s & 20.49 \\
    flower & 0.54 & 17m 00s & 16.63 \\
    horns & 0.81 & 18m 54s & 22.93 \\
    fern & 0.82 & 18m 31s & 15.41 \\
    garden & 4.20 & 16m 23s & 38.94 \\
    kitchen & 1.53 & 7m 25s & 20.06 \\
    train & 1.07 & 8m 24s & 25.69 \\
    truck & 2.59 & 9m 16s & 27.21 \\
    \bottomrule
\end{tabular}
    
    }
    \vspace{-0.05in}
    \label{tab:stylization_resource}
\end{table}
\vspace{-1cm}
\section{Additional Complexity Analysis}
\label{supp:complexity}

Table~\ref{tab:stylization_resource} summarizes the number of Gaussian primitives, the average stylization time, and GPU memory usage across different scenes. The stylization process takes approximately 9–18 minutes per scene depending on scene complexity and rendering configurations. While the stylization time does not show a clear correlation with the number of Gaussians, it can vary depending on scene characteristics and rendering configurations. In contrast, GPU memory usage tends to increase with the number of Gaussian primitives.



\section{User Study Details}
\label{supp:user_study}
\paragraph{Recruitment and procedure.}
 We conducted an online user study with 37 participants to evaluate the perceptual quality of our method. The survey was hosted using an online survey platform, and participants were recruited through our university’s online community and bulletin boards.

Each participant completed evaluation trials comprising four scenes and four style images, resulting in a total of 16 scene–style combinations. In each trial, participants compared the stylized results from our method against four methods: StyleGaussian~\cite{StyleGaussian}, G-Style~\cite{GStyle}, SGSST~\cite{SGSST}, and CLIPGaussian ~\cite{clipgaussian}. All results were presented anonymously and in a randomized order to mitigate potential bias.

\begin{figure*}[p]
    \centering
    \begin{minipage}[d]{0.47\linewidth}
        \vspace{0pt} 
        \centering
        \includegraphics[width=\linewidth, trim=0 3800 0 0, clip]{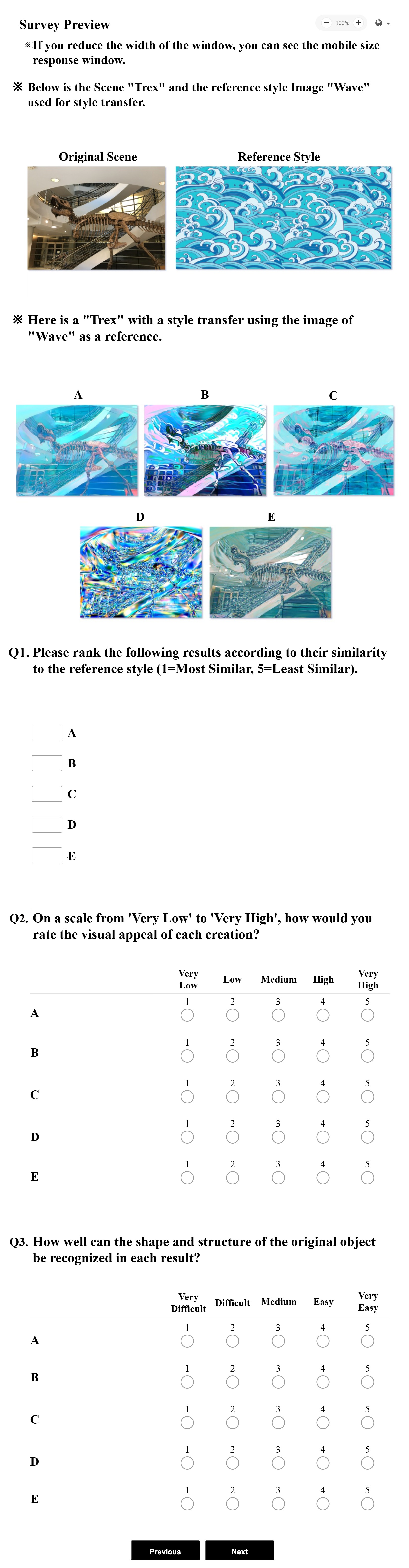}
    \end{minipage}
    \hfill
    \begin{minipage}[c]{0.47\linewidth}
        \vspace{0pt}
        \centering
        \includegraphics[width=\linewidth, trim=0 0 0 2500, clip]{figures/survey.jpeg}
    \end{minipage}
    \caption{A screenshot of the user study interface on the online survey platform.}
    \label{fig:survey_interface}
\end{figure*}

\paragraph{Survey interface.}
Fig.~\ref{fig:survey_interface} illustrates the typical interface presented to participants on the online survey platform. For each trial, participants were provided with the original scene, the target style image, and five anonymized stylized results.
They were then asked to answer the following three questions.

\paragraph{Survey questions.}
\label{ssec:survey_questions} 
Participants were asked the following three questions for each set of results:

\begin{description}
    \item[\textbf{Q1. Style Similarity (Ranking)}]
    ``Please rank the following results according to their similarity to the reference style (1$=$Most Similar, 5$=$Least Similar).''
    \\ \textit{\small Note: This was a ranking question where each method received a unique rank from 1 to 5.}
    
    \item[\textbf{Q2. Visual Appeal (Rating)}]
    ``On a scale from `Very Low' to `Very High', how would you rate the visual appeal of each creation?''
    \\ \textit{\small Note: This was a 5-point Likert scale (1$=$Very Low, 5$=$Very High) applied independently to each result.}
    
    \item[\textbf{Q3. Content Recognizability (Rating)}]
    ``How well can the shape and structure of the original object be recognized in each result?''
    \\ \textit{\small Note: This was a 5-point Likert scale (e.g., 1$=$Very Difficult, 5$=$Very Easy) applied independently to each result.}
    
\end{description}

\paragraph{Aggregated results.}
\label{sec:appendix_results}
 The aggregated results from all 37 participants are presented both 
graphically in the main paper and numerically in this supplement. Fig.~\ref{fig:user_study_2} in the main paper visualizes the detailed distribution of participant responses for each question. For Q1, it depicts the cumulative vote count for each rank, while for Q2 and Q3, it shows the full distribution of 5-point Likert-scale ratings. To provide a concise statistical summary of these distributions, we report the mean scores for all three questions in Table~\ref{tab:user_study}. Together with the distributions shown in Fig.~\ref{fig:user_study_2}, these results demonstrate that our method was consistently preferred by participants, achieving the highest ratings across all evaluation criteria.

\section{Additional Ablation Studies}
\label{supp:additional_ablation}
\begin{figure*}[t] 
\centering 
\includegraphics[width=\linewidth]{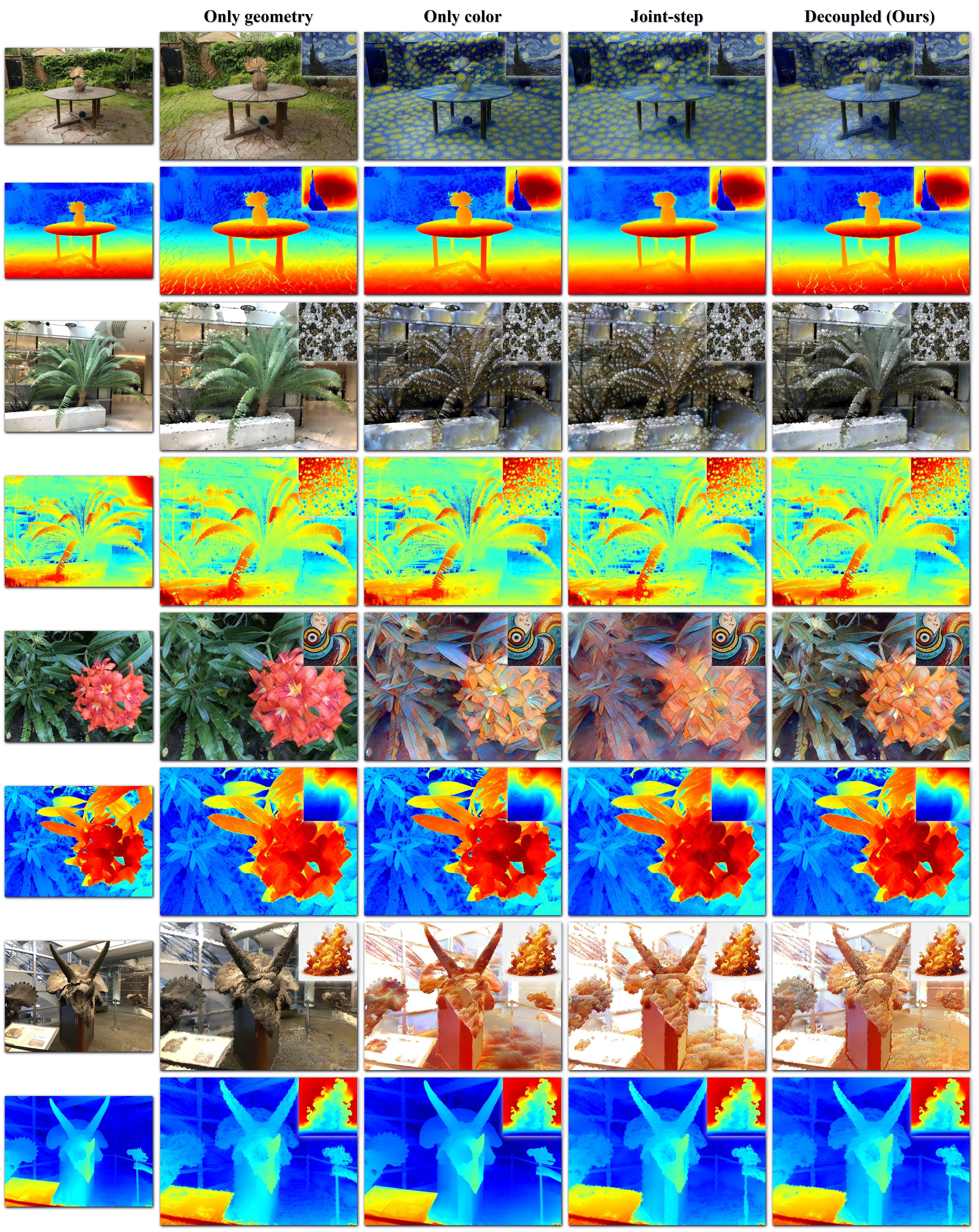} 
\caption{Qualitative comparison of optimization strategies.}
\label{fig:compare_optimization} 
\end{figure*}

\paragraph{Additional ablation study on optimization strategies.}
In Fig.~\ref{fig:compare_optimization},
we compare four optimization strategies for 3D style transfer: only geometry, only color, joint-step, and our decoupled optimization. Only geometry optimization focuses on structural deformation, effectively embedding geometric traits of the target style but lacking color variation, which results in visually incomplete stylization. On the other hand, only color optimization preserves scene geometry but limits style transfer to surface appearance, resulting in weak spatial integration of the style. The joint-step approach updates geometry and color simultaneously, yielding stronger stylization effects but often overfitting to the style appearance. 
Our decoupled optimization achieves a balanced integration of color and geometry, enabling geometry-aware stylization that effectively reflects geometric characteristics of the target style while maintaining appropriate color stylization.

\begin{figure*}[t] 
\centering 
\includegraphics[width=\linewidth]{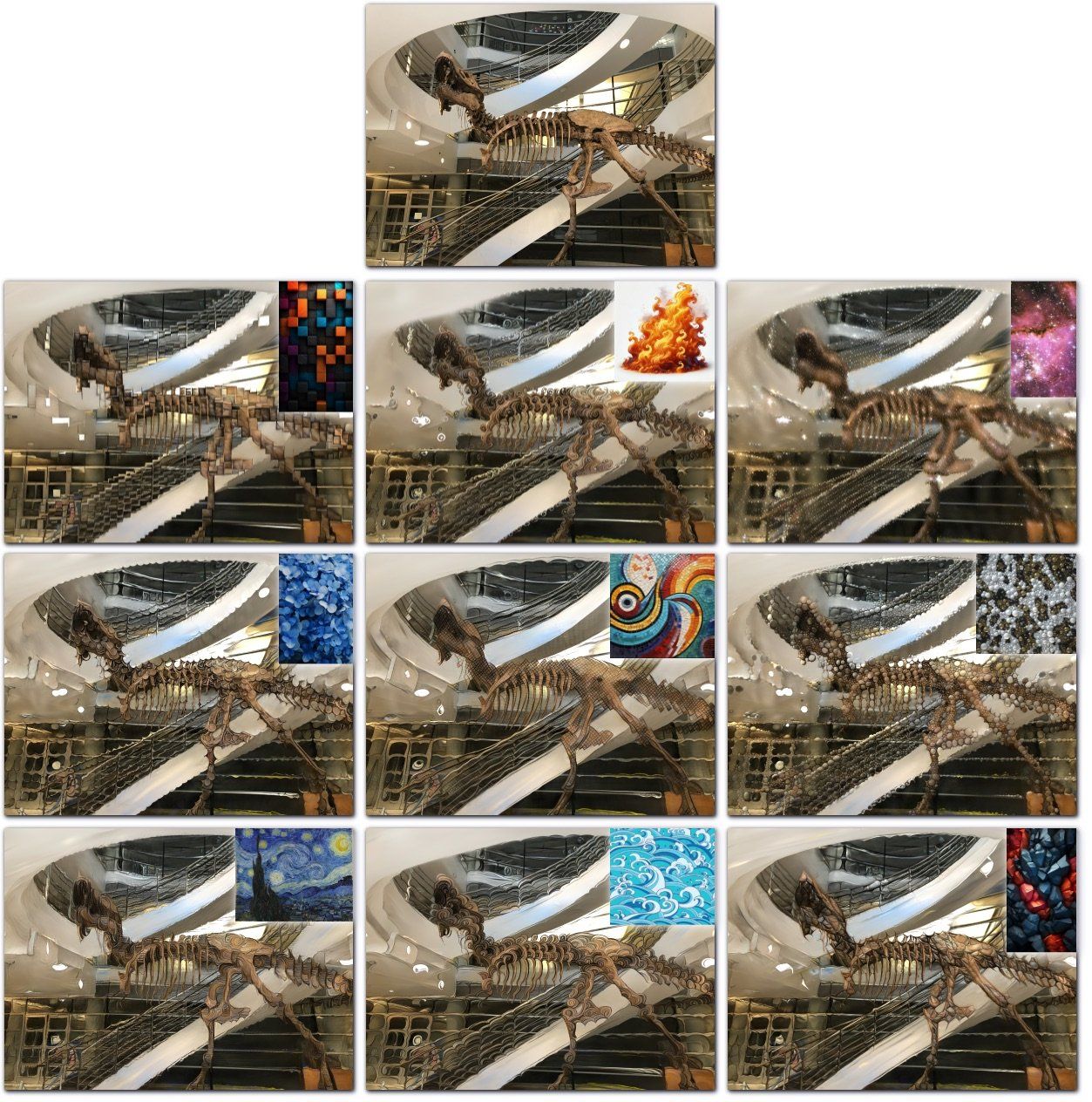} 
\caption{Qualitative results of only geometry optimization. 
}

\label{fig:only_geo_trex} 
\end{figure*}

Fig.~\ref{fig:only_geo_trex} additionally illustrates the results of only geometry optimization. Although it lacks color expressiveness, it still captures structural cues characteristic of the target style, producing distinctive surface reliefs and geometry-aware stylization effects. This observation suggests that geometry-based modulation can serve as a useful component for structural enhancement in 3D style transfer frameworks.

\begin{figure*}[t] 
\centering 
\includegraphics[width=\linewidth]{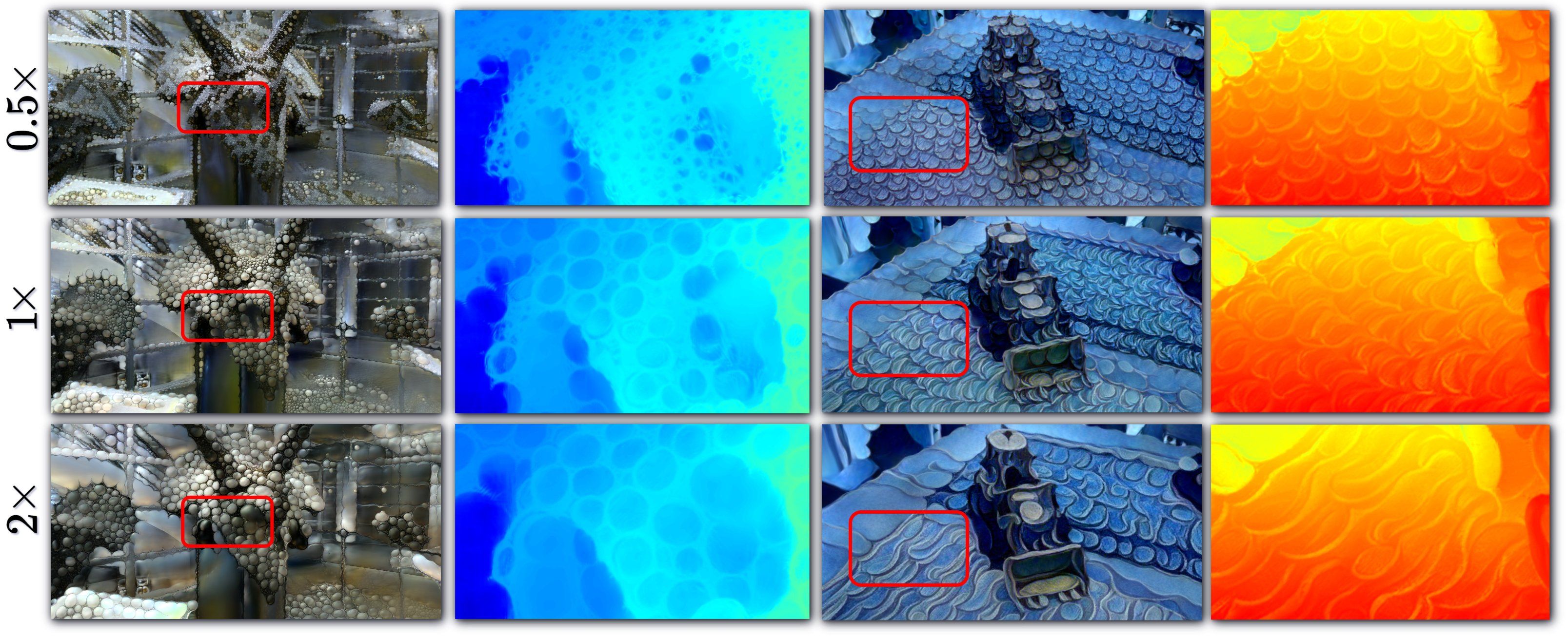} 
\caption{Qualitative results of target texture scale control. 
The scales $0.5\times$, $1\times$, and $2\times$ correspond to target texture resolutions of 
$128 \times 128$, $256 \times 256$, and $512 \times 512$, respectively.
}

\label{fig:ablation_scale} 
\end{figure*}

\paragraph{Ablation study on scale control.}
In Fig.~\ref{fig:ablation_scale}, we investigate the effect of the target texture scale. 
Here, the scale is defined relative to the default target texture resolution of $256 \times 256$: 
$0.5\times$, $1\times$, and $2\times$ correspond to resized resolutions of 
$128 \times 128$, $256 \times 256$, and $512 \times 512$, respectively. 
This parameter controls the strength and granularity of the transferred texture patterns, 
thereby affecting the balance between structural stability and stylistic deformation.

At a smaller scale (\eg, $0.5\times$), the target texture is downsampled to $128 \times 128$, 
which suppresses excessive geometric changes and helps preserve the original scene structure. 
The default scale ($1\times$) provides a balanced result between geometry preservation and texture transfer. 
In contrast, a larger scale (\eg, $2\times$) uses a higher-resolution target texture of $512 \times 512$, 
allowing more pronounced and expressive geometric patterns to appear in the stylized result. 
This qualitative analysis shows that the target texture scale provides a flexible control mechanism 
for adaptively balancing structural fidelity and stylistic expressiveness.

\begin{figure*}[t] 
\centering 
\includegraphics[width=\linewidth]{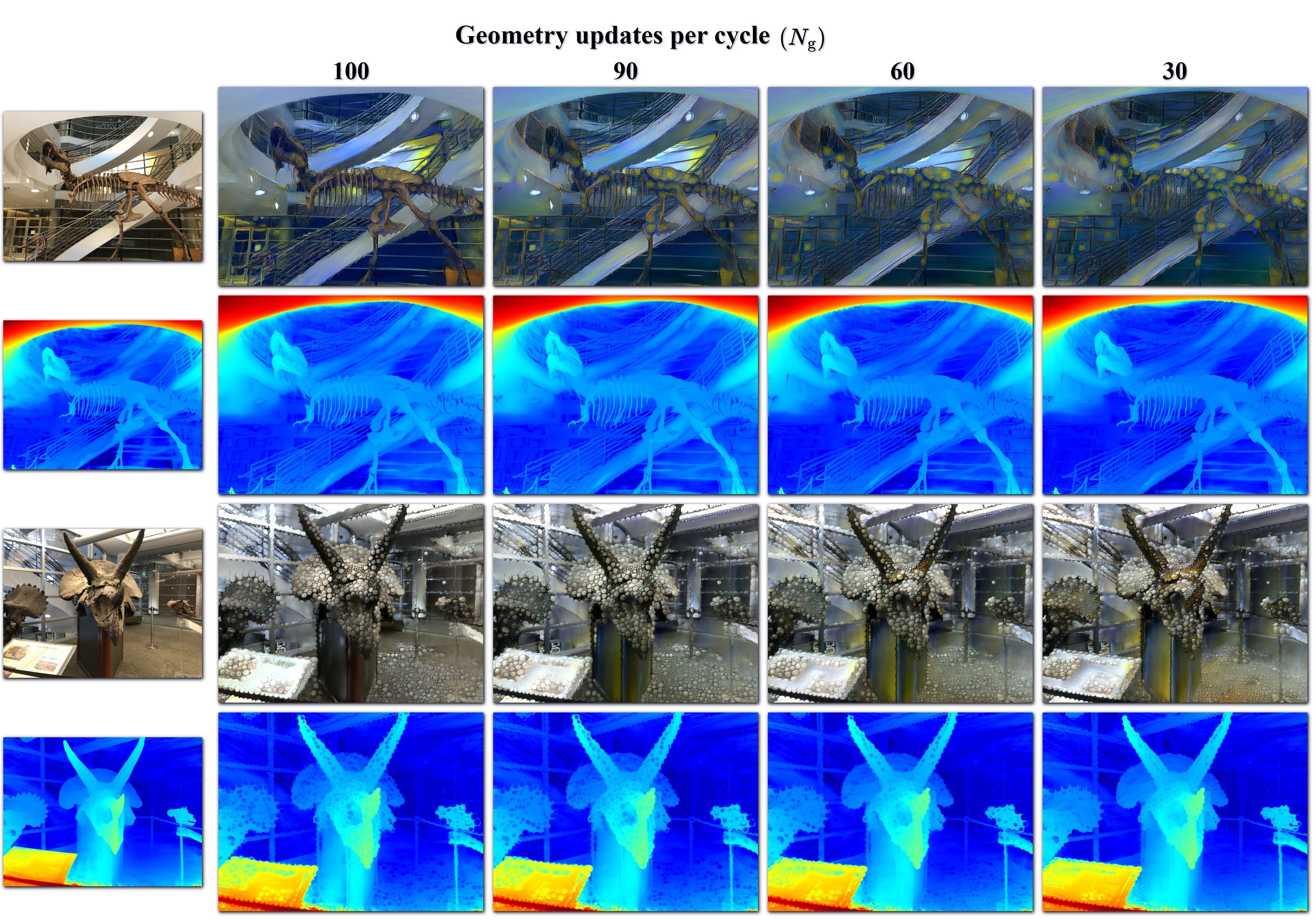} 
\caption{Qualitative results of the ablation study on the geometry/color update ratio.}
\label{fig:ablation_ratio_geo} 
\end{figure*}

\paragraph{Ablation study on the geometry/color update ratio.}
We investigate the impact of varying the ratio between geometry and color updates within each optimization cycle.
Each cycle consists of 100 iterations, where $N_{\mathrm{g}}$ and $N_{\mathrm{c}}$ denote the number of geometry and color update steps, respectively, such that $N_{\mathrm{g}} + N_{\mathrm{c}} = 100$. Table~\ref{tab:optimization_ratio} presents the SIFID scores across different update ratios. 
Our configuration ($N_{\mathrm{g}}=90$, $N_{\mathrm{c}}=10$) achieves the best perceptual score, 
indicating a balanced trade-off between stylization strength and geometric stability. In Fig.~\ref{fig:ablation_ratio_geo}, we present qualitative results showing the effect of varying the geometry/color update ratio. A higher geometry ratio (\eg, $N_{\mathrm{g}}=100$) results in overly rigid structures with limited color adaptation, while a lower ratio (\eg, $N_{\mathrm{g}}=30$) weakens geometric stylization and leads to flatter structures. Our setting ($N_{\mathrm{g}}=90$) yields vivid and spatially consistent stylization, demonstrating that a geometry-dominant update schedule is essential for stable and expressive 3D style transfer.

\begin{table}[t]
\centering
\caption{
    User study results. Average scores of 2,368 responses from 37 participants:
Q1 – \textit{Style Similarity} (↓), Q2 – \textit{Visual Appeal} (↑), and Q3 – \textit{Content Recognizability} (↑).
    }
    \vspace{-0.3cm}
\setlength{\tabcolsep}{6pt}
\resizebox{0.45\columnwidth}{!}{%
\begin{tabular}{ccccc}
    \toprule
    Methods & Q1↓ & Q2↑ & Q3↑ \\
    \midrule
    StyleGaussian~\cite{StyleGaussian} & 4.78 & 1.95 & 1.90 \\
    G-Style~\cite{GStyle} & 2.64 & 3.65 & 3.83 \\
    SGSST~\cite{SGSST} & \underline{2.26} & \underline{3.74} & 3.70 \\
    CLIPGaussian~\cite{clipgaussian} & 3.55 & 3.45 & \underline{4.19} \\
    Ours & \textbf{1.75} & \textbf{4.07} & \textbf{4.32} \\
    \bottomrule
\end{tabular}
    
    }
    \vspace{-0.05in}
    \label{tab:user_study}
\end{table}

\begin{table}[t]
\centering
\caption{Ablation study on the geometry/color update ratio. Comparison of SIFID scores across different $N_{\mathrm{g}}$/$N_{\mathrm{c}}$ values.}
\setlength{\tabcolsep}{6pt}
\vspace{-0.2cm}
\resizebox{0.6\columnwidth}{!}{%
\begin{tabular}{ccccccc}
    \toprule
    $N_{\mathrm{g}}$ & 0 & 30 & 60 & 90 (ours) & 100 \\
    \midrule
    SIFID↓ & 1.5244 & 1.3272 & 1.2526 & \textbf{1.1736} & \underline{1.1885} \\
    \bottomrule
\end{tabular}
    }

    \vspace{-0.05in}
    \label{tab:optimization_ratio}
\end{table}

\paragraph{Ablation study on GCFM hyperparameter.}
In Table~\ref{tab:lambda_gcfm}, we evaluate the sensitivity of our framework to the weighting factor $\lambda_{\mathrm{GC}}$ in GCFM. The SIFID scores remain largely consistent across a wide range of values, indicating that GCFM is robust to moderate variations in its weighting.
When the term is not used (\ie, $\lambda_{\text{GC}}=0$), the performance drops notably, suggesting that the GCFM plays a crucial role in guiding consistent stylization.
Across a wide range of $\lambda_{\mathrm{GC}}$ values, our method produces stable and visually coherent stylization results. We adopt $\lambda_{\mathrm{GC}} = 2.0$ in all experiments, as it consistently provides visually appealing and structurally coherent results across diverse scenes. This result demonstrates that our method is robust to variations in $\lambda_{\mathrm{GC}}$, producing consistent stylization quality with stable optimization behavior.

\begin{table}[t]
\centering
\caption{Ablation study on GCFM hyperparameter. SIFID remains stable across various weights, confirming that our method is robust to the choice of $\lambda_{\text{GC}}$.}
\vspace{-0.1cm}
\setlength{\tabcolsep}{6pt}
\resizebox{0.55\columnwidth}{!}{%
\begin{tabular}{ccccccc}
    \toprule
    $\lambda_{\text{GC}}$ & 0.0 & 1.0 & 2.0 (ours) & 5.0 & 10.0 \\
    \midrule
    SIFID↓ & 1.5718 & \textbf{1.1683} & \underline{1.1736} & 1.1841 & 1.1878 \\
    \bottomrule
\end{tabular}
    }
    \vspace{-0.05in}
    \label{tab:lambda_gcfm}
\end{table}

\section{Comparison with StylizedGS}
\label{supp:stylizedGS}
\begin{figure*}[t] 
\centering 
\includegraphics[width=\linewidth]{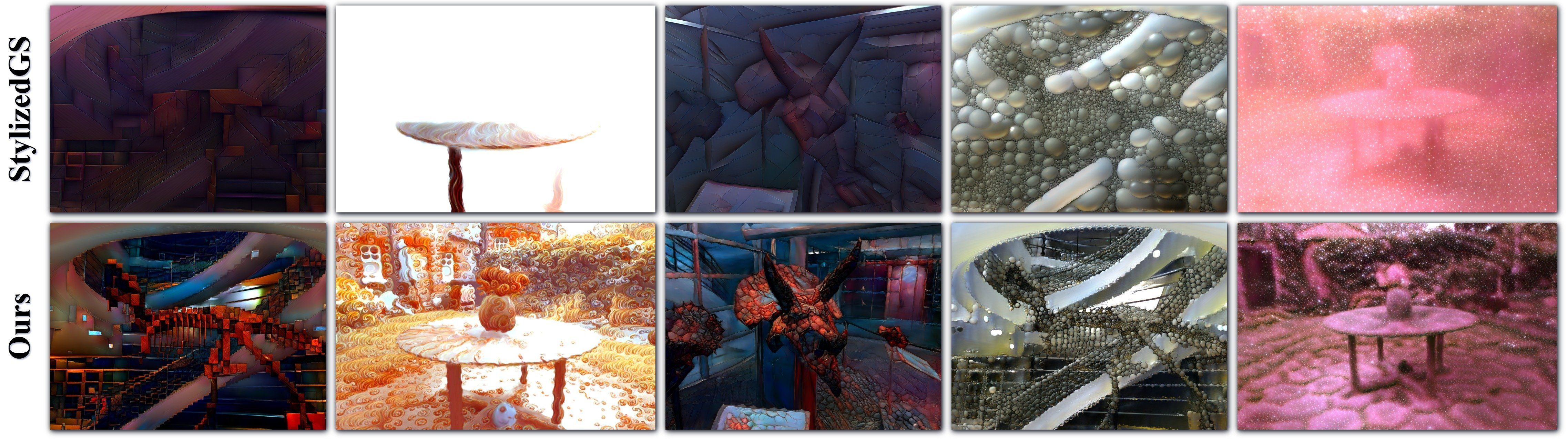} 
\caption{Qualitative comparison with StylizedGS.}
\label{fig:compare_stylizedgs} 
\end{figure*}

In Fig.~\ref{fig:compare_stylizedgs}, we qualitatively compare our method with StylizedGS~\cite{StylizedGS}.
While StylizedGS achieves higher quantitative consistency (LPIPS and RMSE) under both short- and long-range evaluations, this can be attributed to its overly smoothed and desaturated renderings rather than genuine structural coherence. As illustrated in the figure, StylizedGS tends to lose geometric details and fails to express distinctive style characteristics, yielding relatively flat and texture-suppressed outputs with diminished 3D structure. In contrast, our method preserves spatial geometry and reproduces vivid, structurally consistent stylization patterns. These results suggest that the high consistency scores of StylizedGS may not directly reflect superior visual quality, but rather a loss of stylization richness—whereas our framework achieves a balance between geometric coherence and expressive style fidelity.



\section{Additional Qualitative Results}
\label{supp:additional_qualitative}
\begin{figure*}[t] 
\centering 
\includegraphics[width=\linewidth]{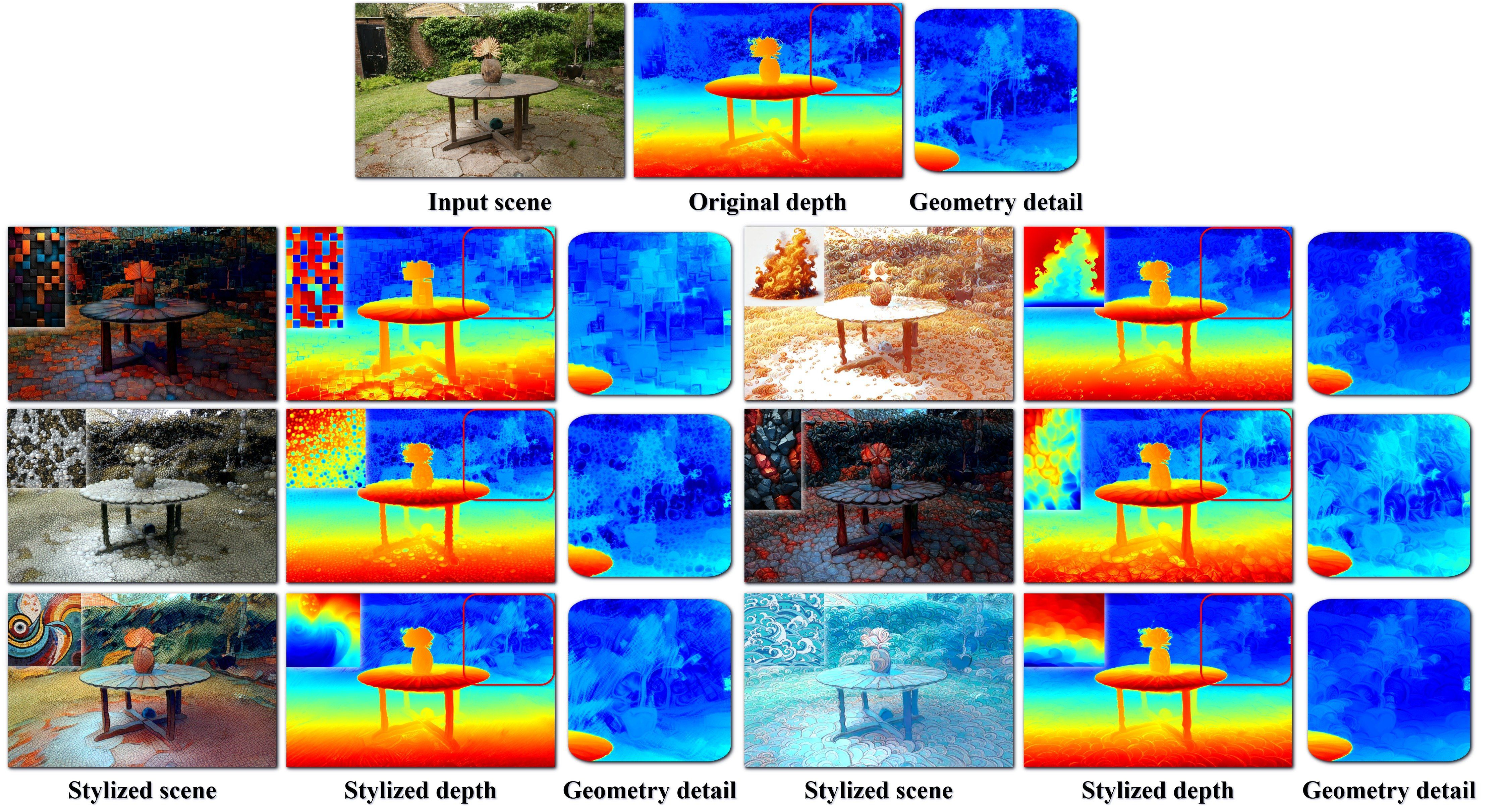} 
\caption{Stylization results for the \textit{Garden} scene, including corresponding depth maps.}
\label{fig:garden_style} 
\end{figure*}

\begin{figure*}[t] 
\centering 
\includegraphics[width=\linewidth]{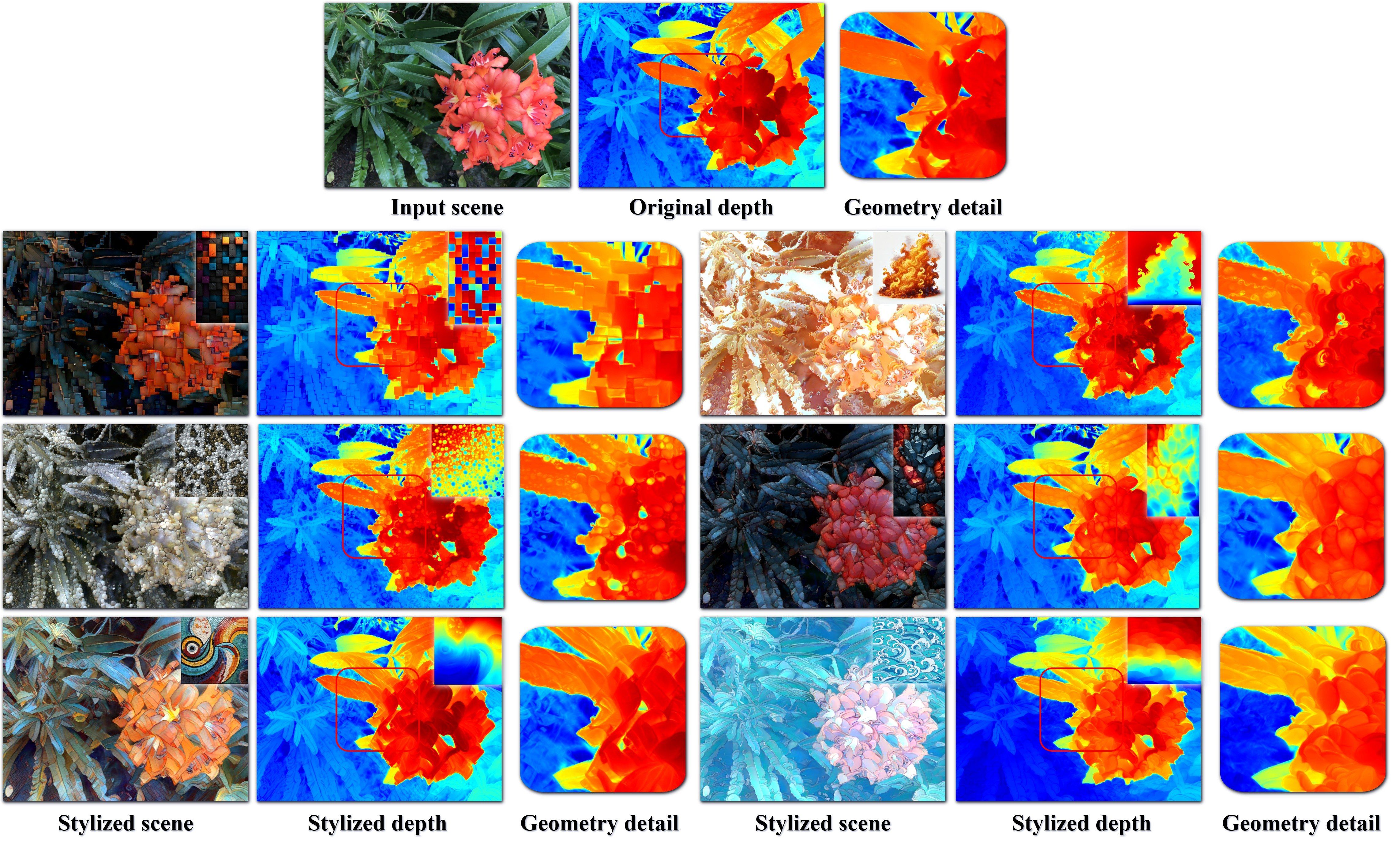} 
\caption{Stylization results for the \textit{Flower} scene, including corresponding depth maps.} 
\label{fig:flower_style} 
\end{figure*}

\begin{figure*}[t] 
\centering 
\includegraphics[width=0.93\linewidth]{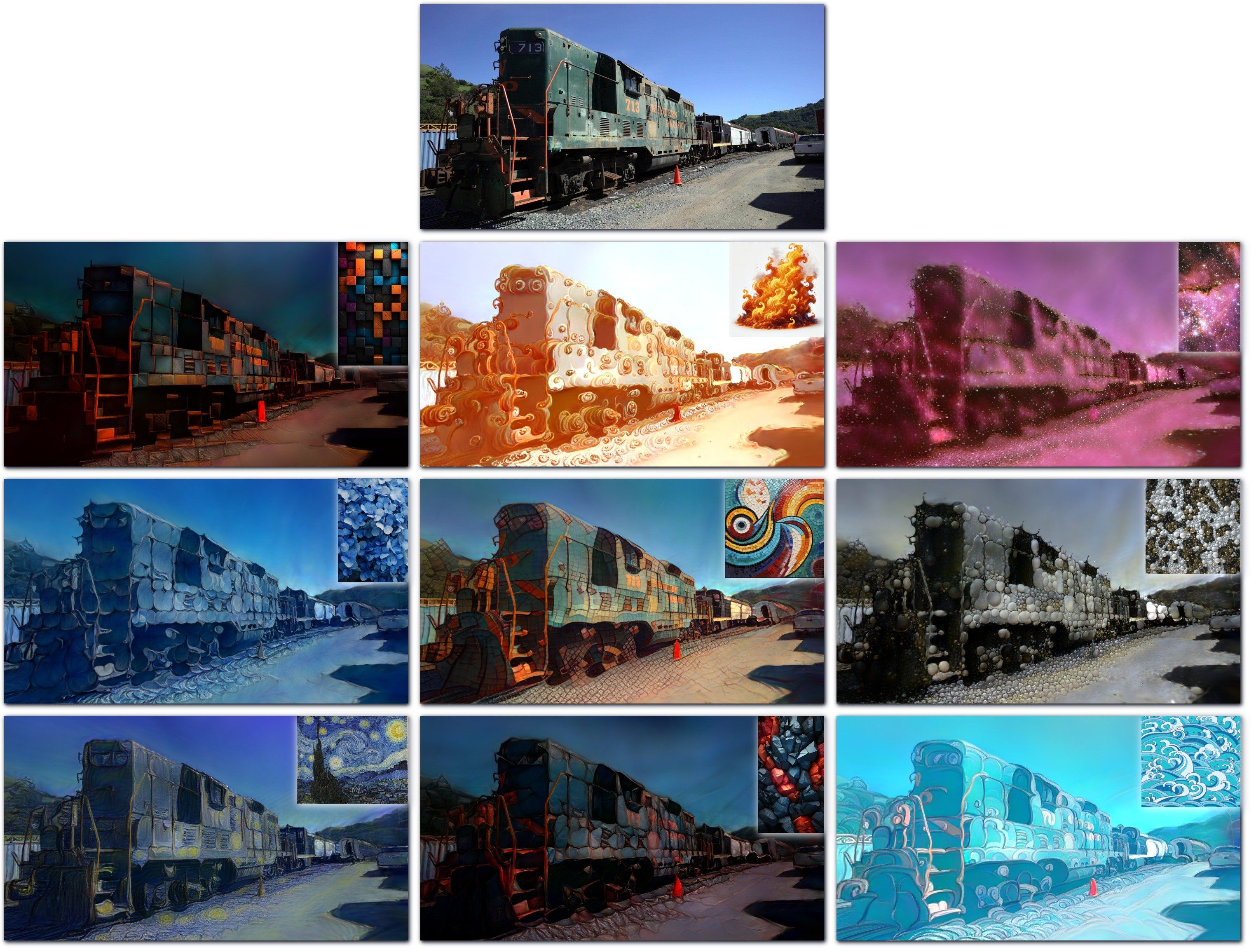} 
\caption{\textit{Train} scene stylization results with nine exemplar styles.}
\label{fig:train_style} 
\end{figure*}

\begin{figure*}[t] 
\centering
\includegraphics[width=0.93\linewidth]{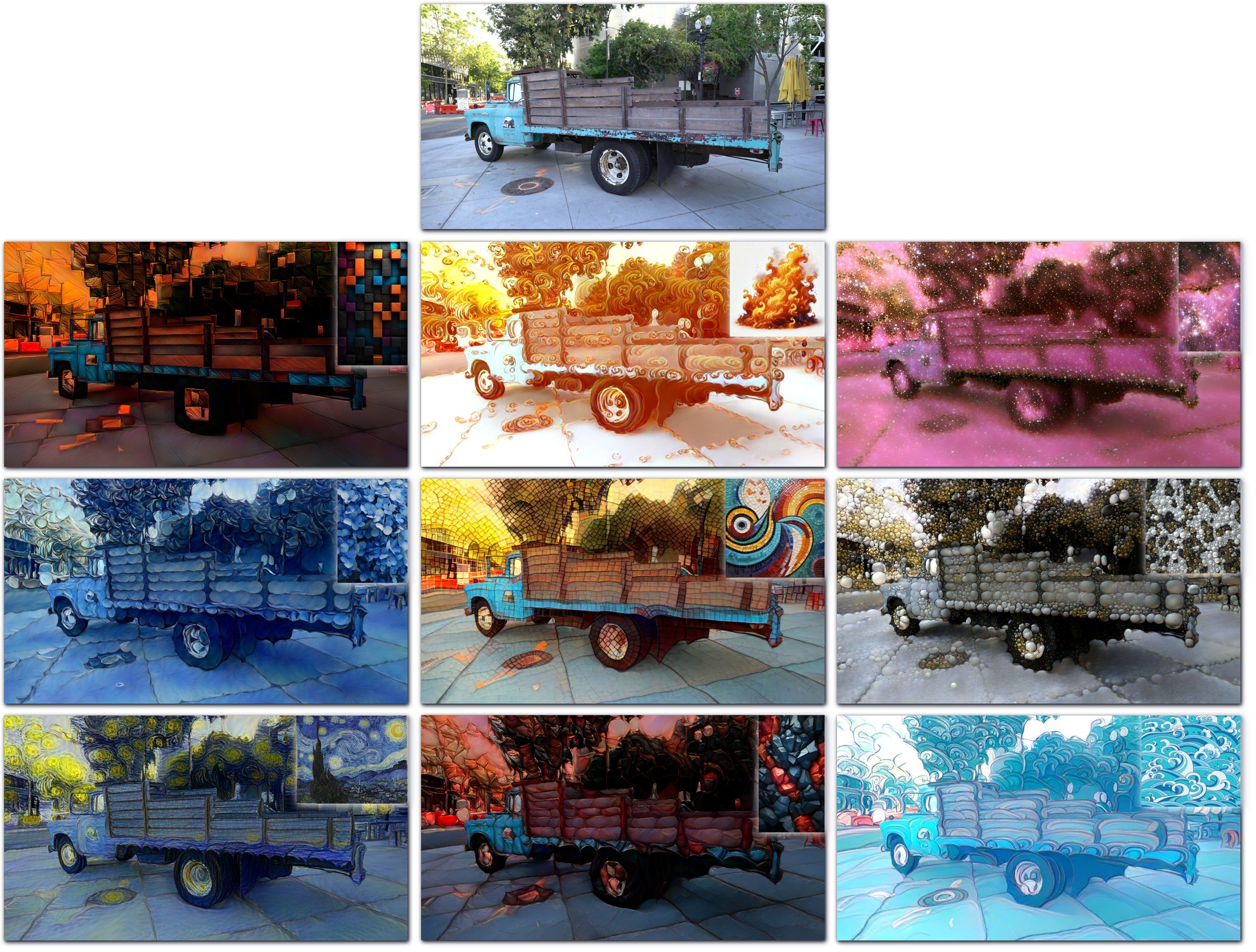} 
\caption{\textit{Truck} scene stylization results with nine exemplar styles.}
\label{fig:truck_style} 
\end{figure*}

\begin{figure*}[t] 
\centering 
\includegraphics[width=0.75\linewidth]{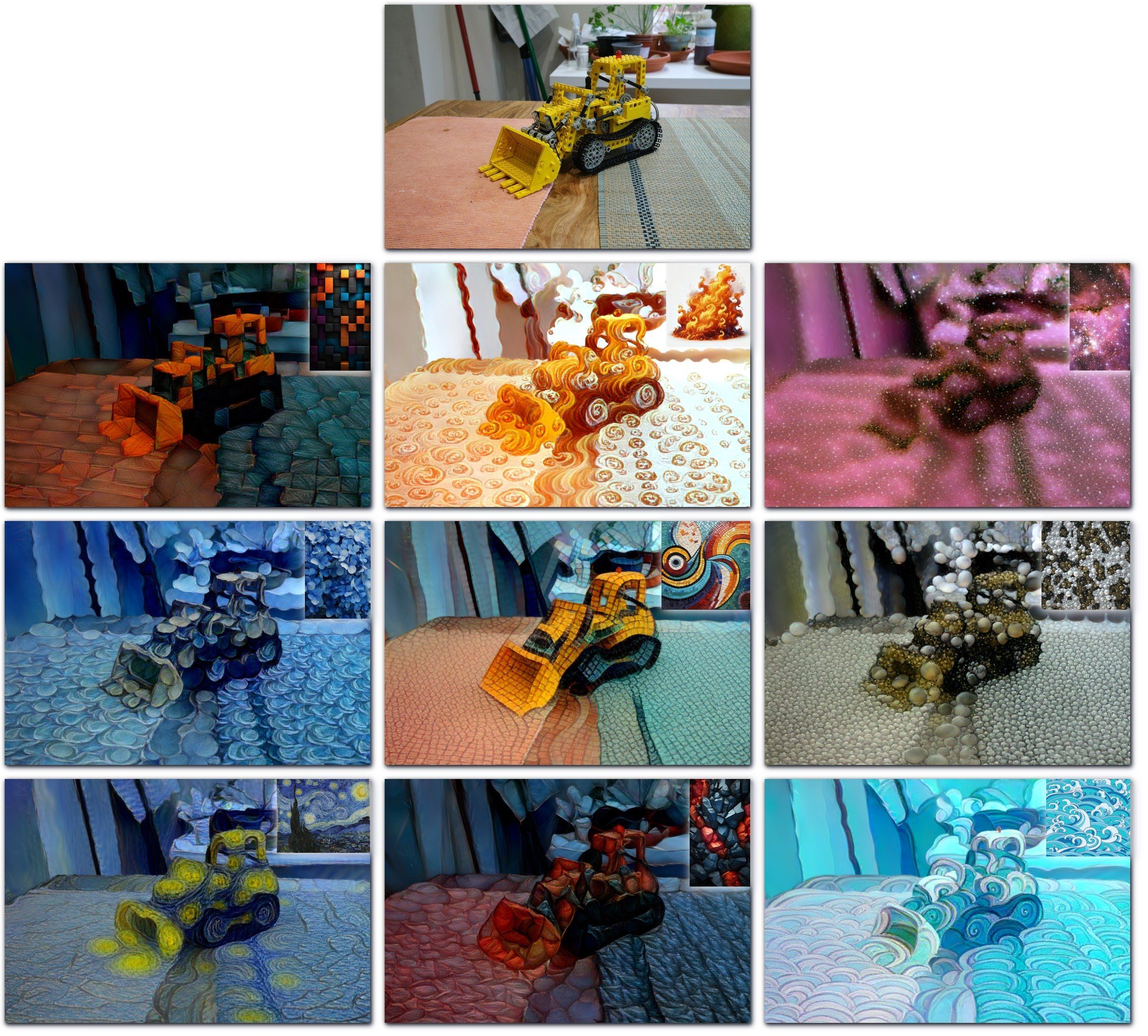} 
\caption{\textit{Kitchen} scene stylization results with nine exemplar styles.}
\label{fig:kitchen_style} 
\end{figure*}

\begin{figure*}[t] 
\centering 
\includegraphics[width=0.75\linewidth]{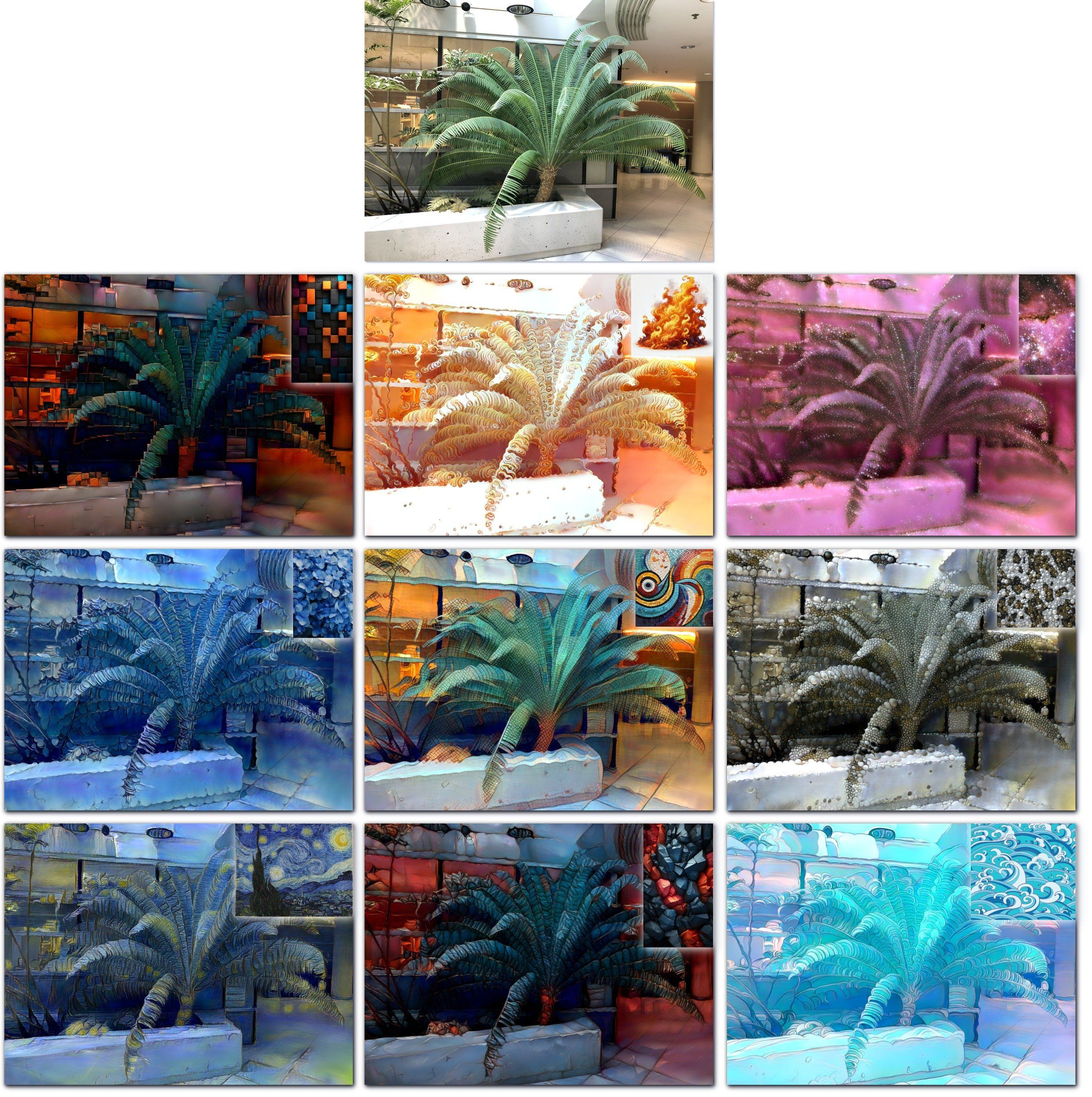} 
\caption{\textit{Fern} scene stylization results with nine exemplar styles.}
\label{fig:fern_style} 
\end{figure*}

\begin{figure*}[t] 
\centering 
\includegraphics[width=\linewidth]{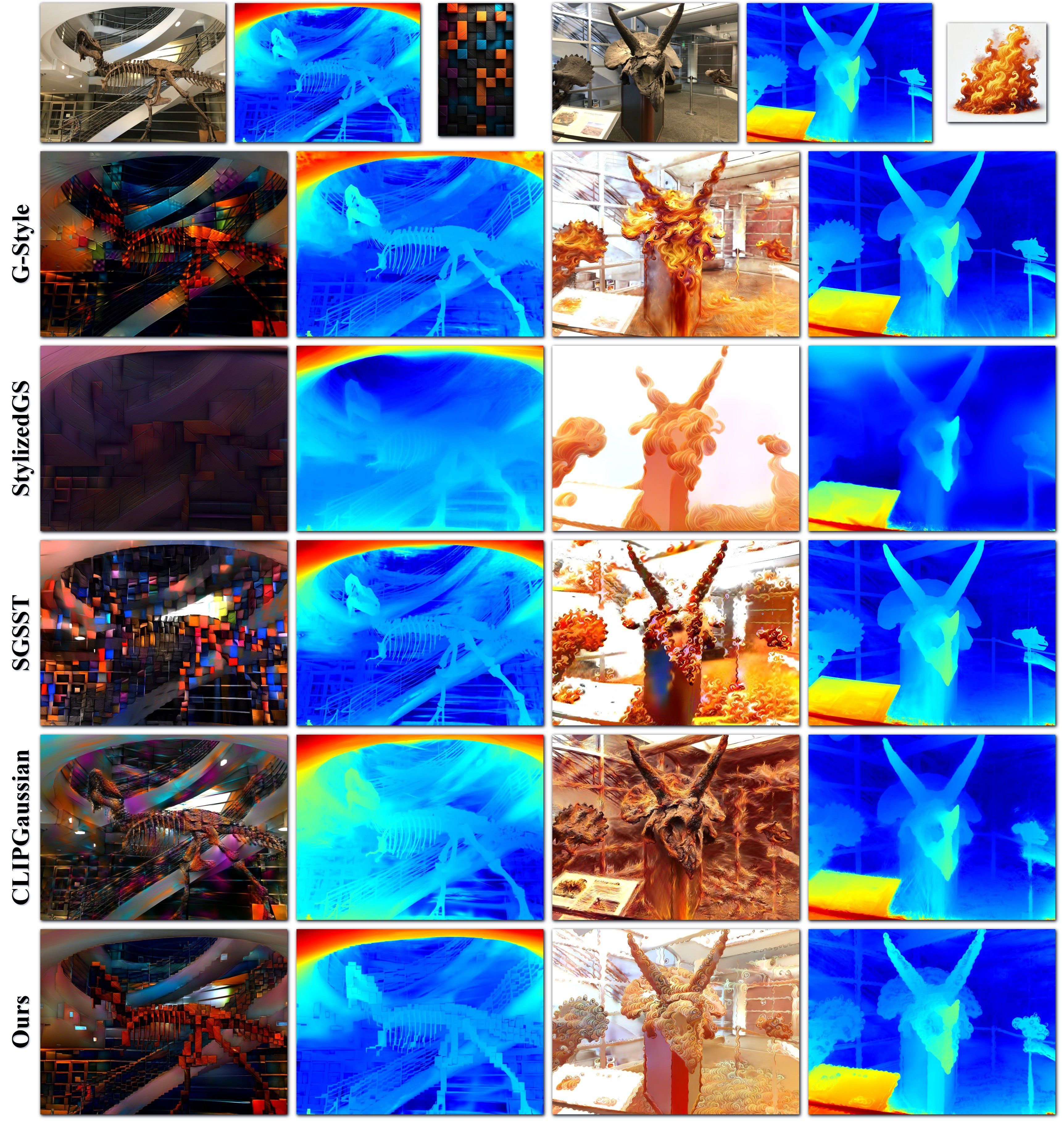} 
\caption{Qualitative comparison of the \textit{Trex} and \textit{Horns} scenes.
}
\label{fig:compare_results} 
\end{figure*}

\begin{figure*}[t] 
\centering 
\includegraphics[width=\linewidth]{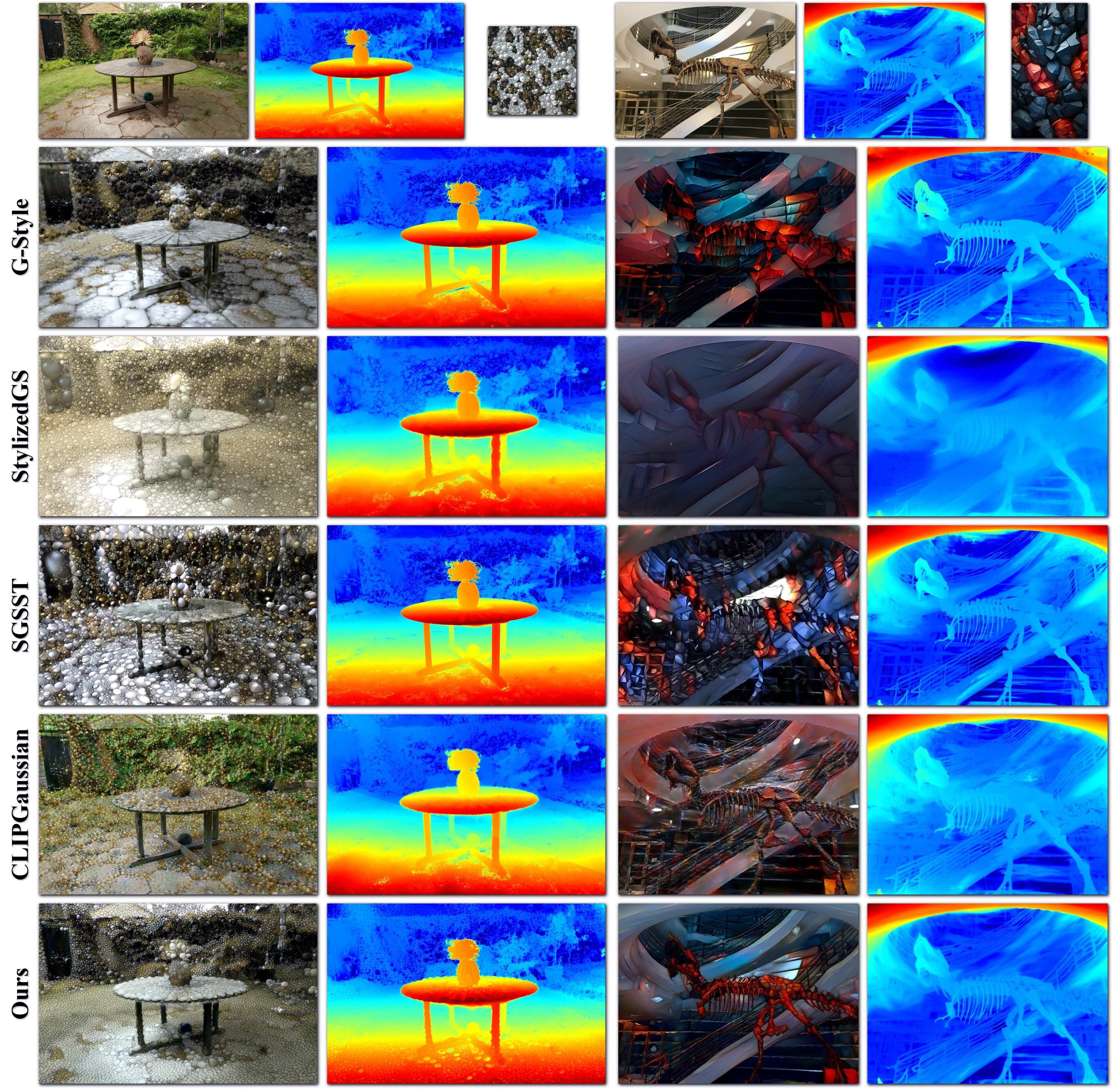} 
\caption{Qualitative comparison of the \textit{Garden} and \textit{Trex} scenes.}
\label{fig:compare_results3} 
\end{figure*}

\begin{figure*}[t] 
\centering 
\includegraphics[width=\linewidth]{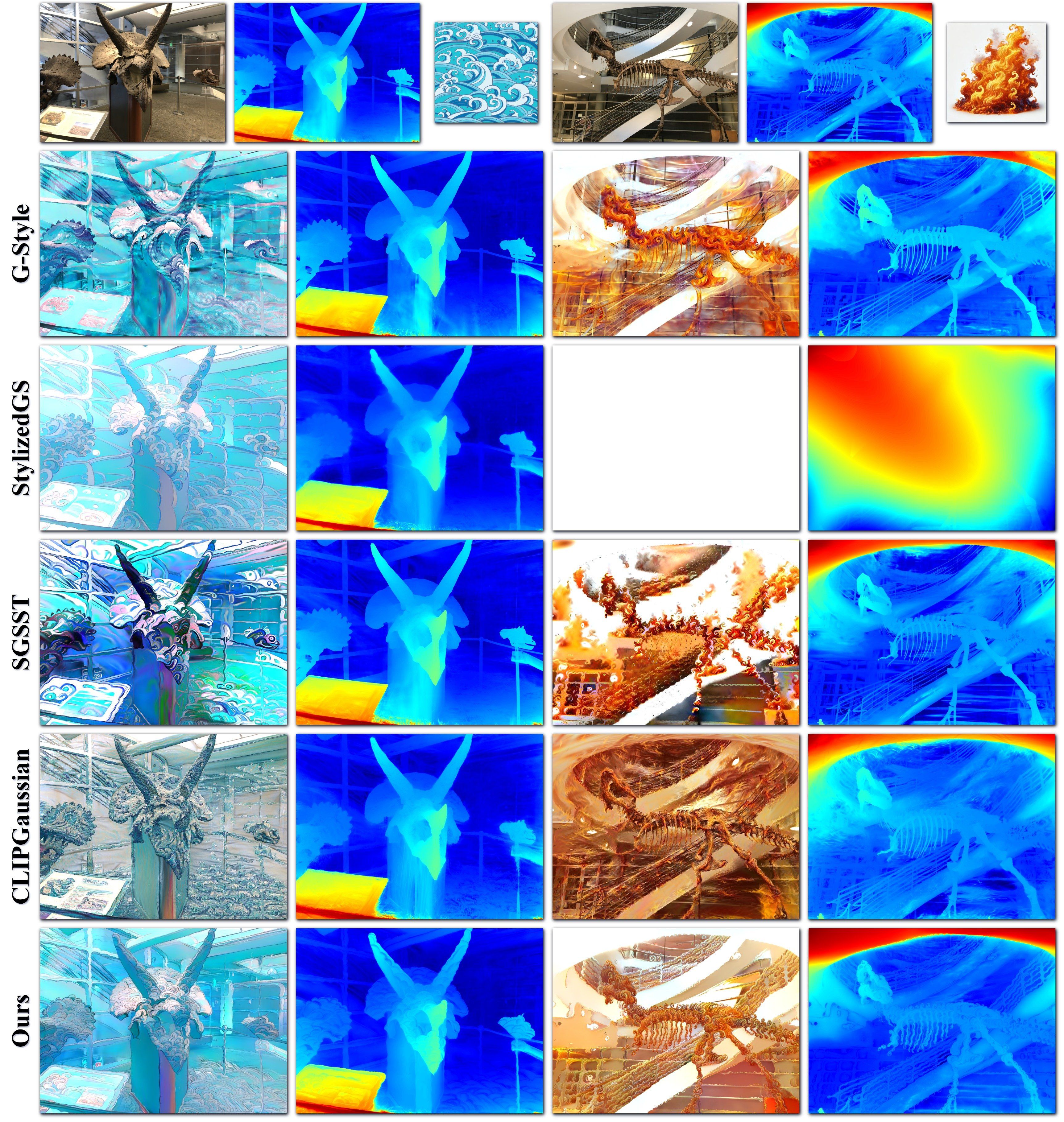} 
\caption{Qualitative comparison of the \textit{Horns} and \textit{Trex} scenes.}
\label{fig:compare_results4} 
\end{figure*}

\begin{figure*}[t] 
\centering 
\includegraphics[width=\linewidth]{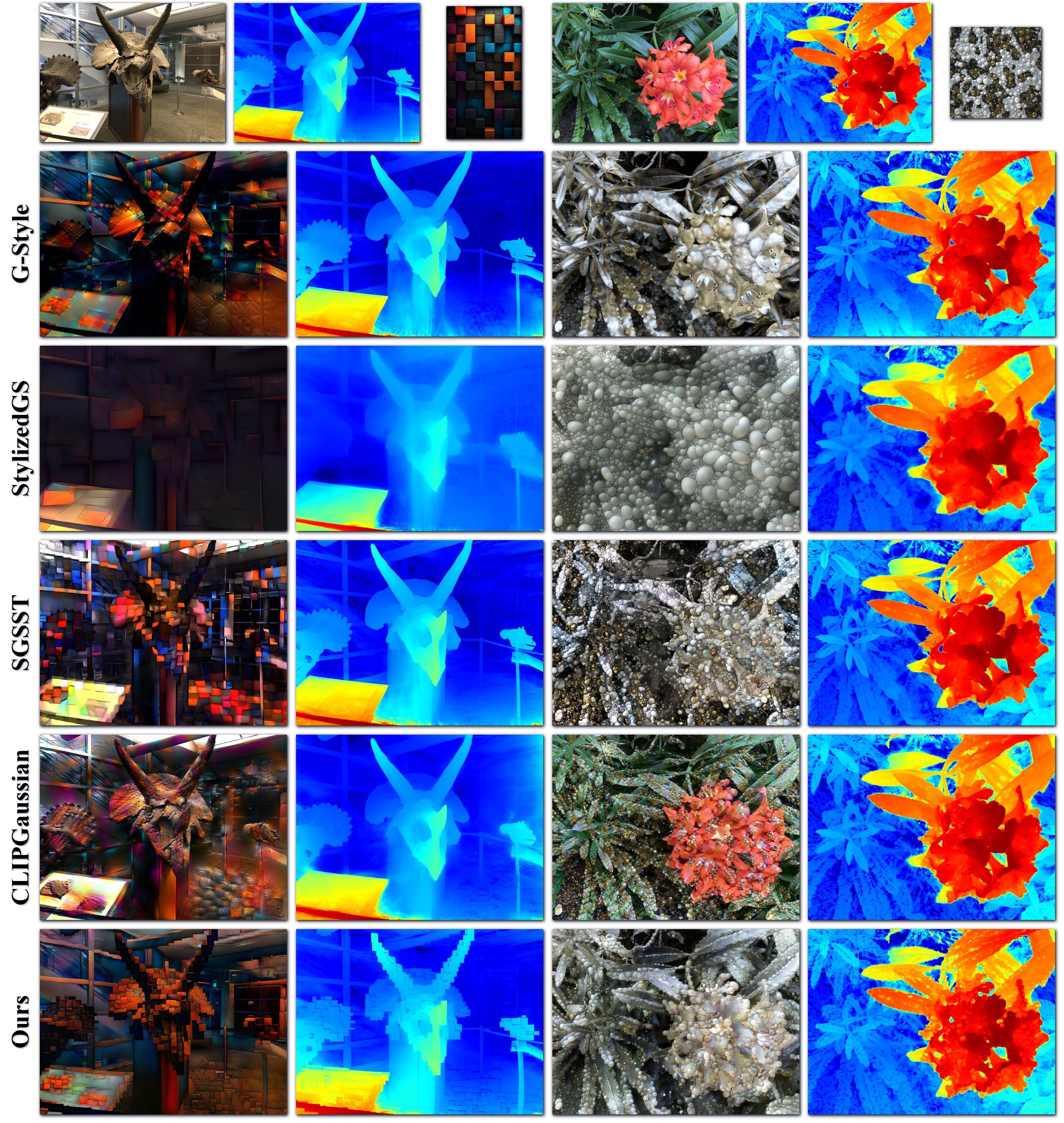} 
\caption{Qualitative comparison of the \textit{Horns} and \textit{Flower} scenes.}
\label{fig:compare_results5} 
\end{figure*}

This section presents additional qualitative stylization results and comparative analyses.
\paragraph{Scene-wise stylization results.}
Figs.~\ref{fig:garden_style}–\ref{fig:fern_style} show stylization outcomes across multiple 3D scenes, including \textit{Garden}, \textit{Flower}, \textit{Train}, \textit{Truck}, \textit{Kitchen}, and \textit{Fern}.
Figs.~\ref{fig:garden_style} and \ref{fig:flower_style} present stylized results along with the corresponding depth maps, demonstrating that the geometric structure is effectively adapted to the target style while preserving the original scene content.
Figs.~\ref{fig:train_style}-\ref{fig:fern_style} further illustrate stylizations using all exemplar styles adopted in our framework, showing consistent geometry–appearance adaptation across diverse environments.
Overall, our method successfully preserves the structural integrity of the scene while transferring both color and geometric characteristics of the target style.
\paragraph{Comparisons with other methods.}
Figs.~\ref{fig:compare_results}–\ref{fig:compare_results5} provide additional comparisons with existing stylization approaches across diverse scenes and styles.
Even under varying scene layouts and style characteristics, our method yields more stable and expressive stylizations than competing methods, maintaining geometric clarity and spatially coherent style patterns throughout the 3D space.
These results complement the quantitative analyses in the main paper and further validate the robustness and visual fidelity of our proposed framework.

\end{document}